\documentclass[10pt,twocolumn,letterpaper]{article}

\usepackage{cvpr}              % To produce the CAMERA-READY version
\usepackage[dvipsnames]{xcolor}

\definecolor{cvprblue}{rgb}{0.21,0.49,0.74}
\usepackage[pagebackref,breaklinks,colorlinks,allcolors=cvprblue]{hyperref}

\usepackage{booktabs}
\usepackage{amsmath,amssymb} % define this before the line numbering.

\usepackage{sg-ibs-macros}
\usepackage{multirow}
\usepackage{pifont}
\newcommand{\cmark}{\ding{51}}
\newcommand{\xmark}{\ding{55}}

\def\paperID{11} % *** Enter the Paper ID here
\def\confName{CVPR}
\def\confYear{2025}

\title{PatchContrast: Self-Supervised Pre-Training for 3D Object Detection}

\author{
  \begin{tabular}{cc}
    Oren Shrout & Ori Nizan \\
    Technion, Israel & Technion, Israel \\
    {\tt\small shrout.oren@campus.technion.ac.il} & {\tt\small snizori@campus.technion.ac.il}
    \\
    Yizhak Ben-Shabat & Ayellet Tal \\
    Technion, Israel & Technion, Israel \\
    {\tt\small sitzikbs@technion.ac.il} & {\tt\small ayellet@ee.technion.ac.il}
  \end{tabular}
}

\begin{document}
\maketitle

\begin{abstract}
    Accurately detecting objects in the environment is a key challenge for autonomous vehicles.
    However, obtaining annotated data for detection is expensive and time-consuming.
    We introduce \textit{PatchContrast}, a novel self-supervised point cloud pre-training framework for 3D object detection.
    We propose to utilize two levels of abstraction to learn discriminative representation from unlabeled data: proposal-level and patch-level.
    The proposal-level aims at localizing objects in relation to their surroundings, whereas the patch-level adds information about the internal connections between the object's components, hence distinguishing between different objects based on their individual components.
    We demonstrate how these levels can be integrated into self-supervised pre-training for various backbones to enhance the downstream 3D detection task.
    We show that our method outperforms existing state-of-the-art models on three commonly-used 3D detection datasets.
\end{abstract}

%%%%%%%%%%%%%%%%%%%%%%%%%%%%%%%%%
% section: Introduction
%%%%%%%%%%%%%%%%%%%%%%%%%%%%%%%%%
\section{Introduction}
% paragraph: motivation
One of the key challenges in autonomous driving is the accurate detection---localization and classification---of objects in the environment, e.g. \textit{Pedestrians}, \textit{Vehicles}, and \textit{Cyclists}. 
To overcome this challenge, autonomous vehicles are equipped with 3D LiDAR scanners that produce 3D point cloud data, which provides detailed information about the surrounding environment.
% Current state-of-the-art approaches for object detection rely on learning an effective embedding that can capture complex features \cite{yin2021center,pvrcnn, pvrcnn++, pointrcnn}. 
% These approaches greatly rely on having annotated data, in the form of 3D bounding boxes.
% However, annotating 3D point cloud data is expensive and time-consuming.
% For example, a typical data acquisition vehicle can collect about $25K$ 3D point cloud frames within an hour, but a skilled annotator can only annotate up to $25$ frames per hour~\cite{once}.
State-of-the-art object detection methods learn effective embeddings to capture complex features~\cite{yin2021center,pvrcnn,pvrcnn++,pointrcnn}, but they heavily depend on annotated 3D bounding boxes.
However, annotating 3D point cloud data is expensive and time-consuming.
For instance, a vehicle can collect $25\mathrm{K}$ frames per hour, while a skilled annotator labels only about $25$~\cite{once}.

Our objective is to learn a discriminative representation of objects for 3D object detection, even in the absence of labeled data.
The primary challenge is how to acquire representations that can be learned effectively without relying on annotations.
By addressing this challenge, we will be able to leverage the available vast amount of unlabeled data.

To address this challenge, we propose utilizing {\em Self-Supervised Learning (SSL)}, which has emerged as a promising approach for learning embeddings from unlabeled data.  
Recent works~\cite{pointcontrast,depthcontrast,GCC-3D,STRL} have presented various self-supervised frameworks for 3D detection. 
The common thread is to pre-train a detector's backbone on a large unlabeled dataset using contrastive learning and then use a small annotated dataset to learn detection-specific features.  
This approach has demonstrated improved performance compared to supervised learning with limited data trained from scratch. 
A key difference between the above methods is the level of abstraction used, either a global scene representation~\cite{depthcontrast,STRL} or a local point/voxel representation~\cite{pointcontrast,GCC-3D}.
Scene-level representation might fail to capture fine details while only point/voxel-level might miss the broader context required to classify objects.
To address this issue and improve localization, ~\cite{proposalcontrast} propose an intermediate level, the {\em proposal level,} which is a subset of points likely to capture an object within its surroundings.

\begin{figure*}[t]
    \centering
    \begin{tabular}{ccc}
        \includegraphics[width=.24\linewidth]{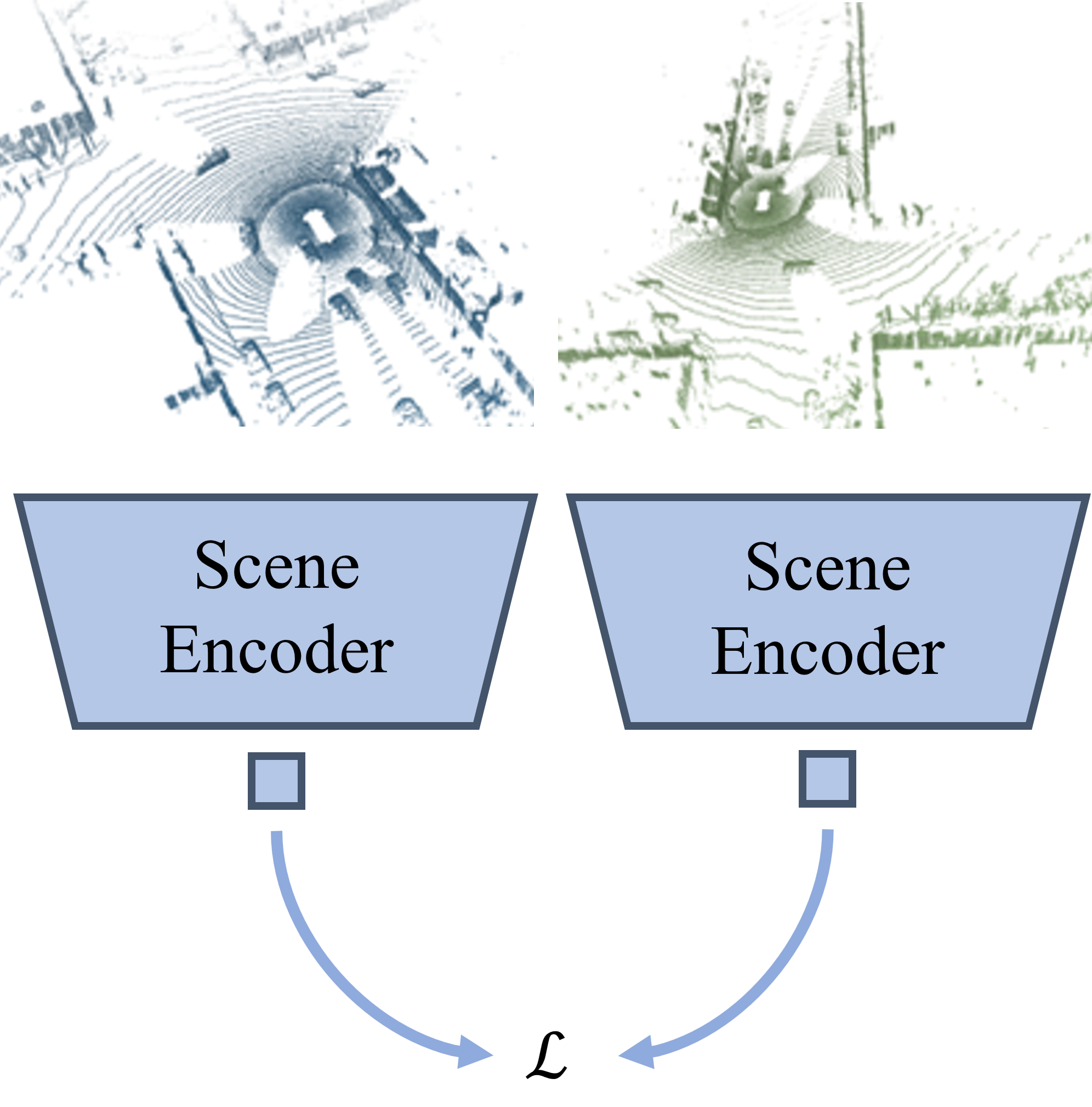} &
        \includegraphics[width=.24\linewidth]{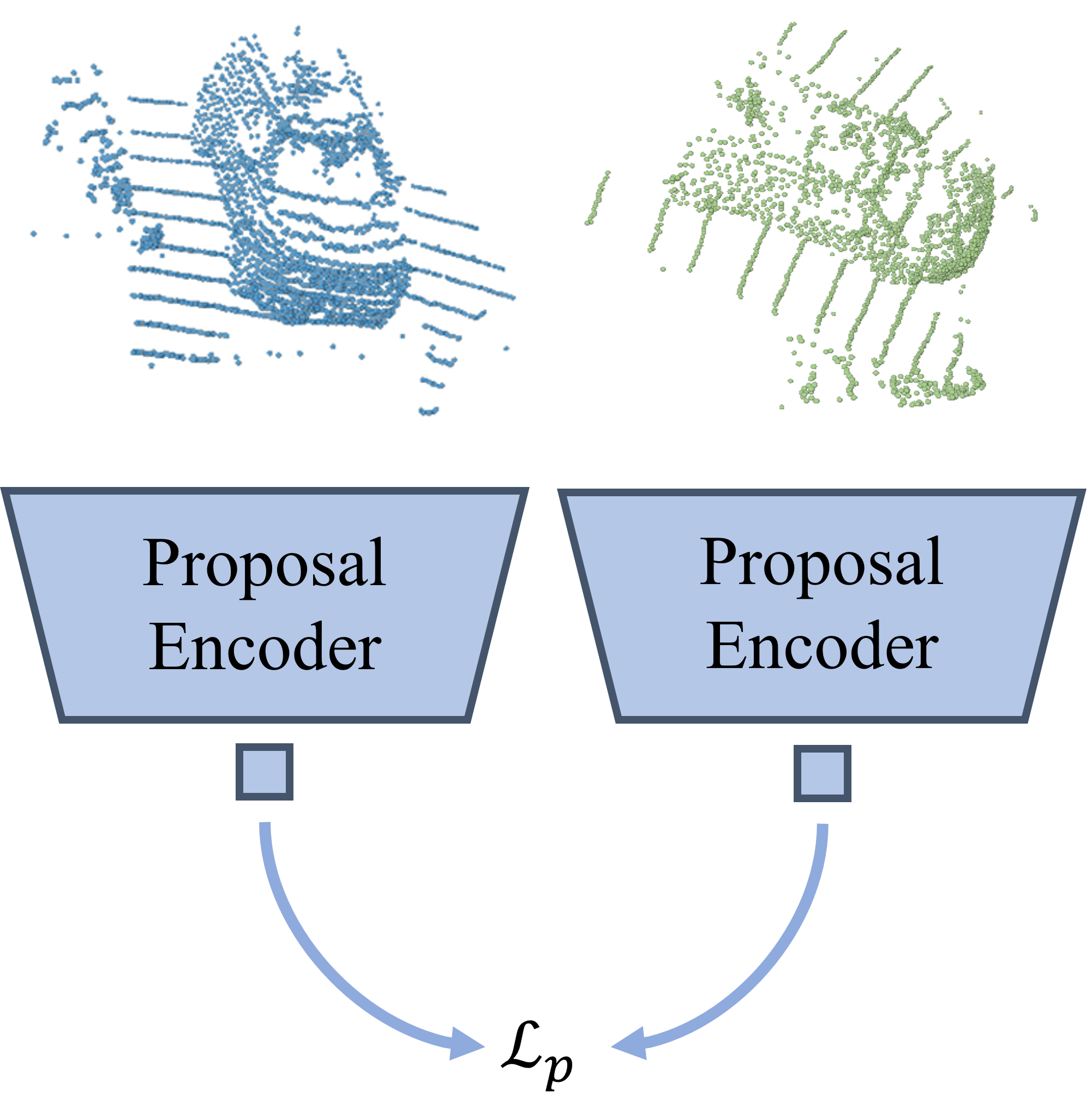} &
        \includegraphics[width=.48\linewidth]{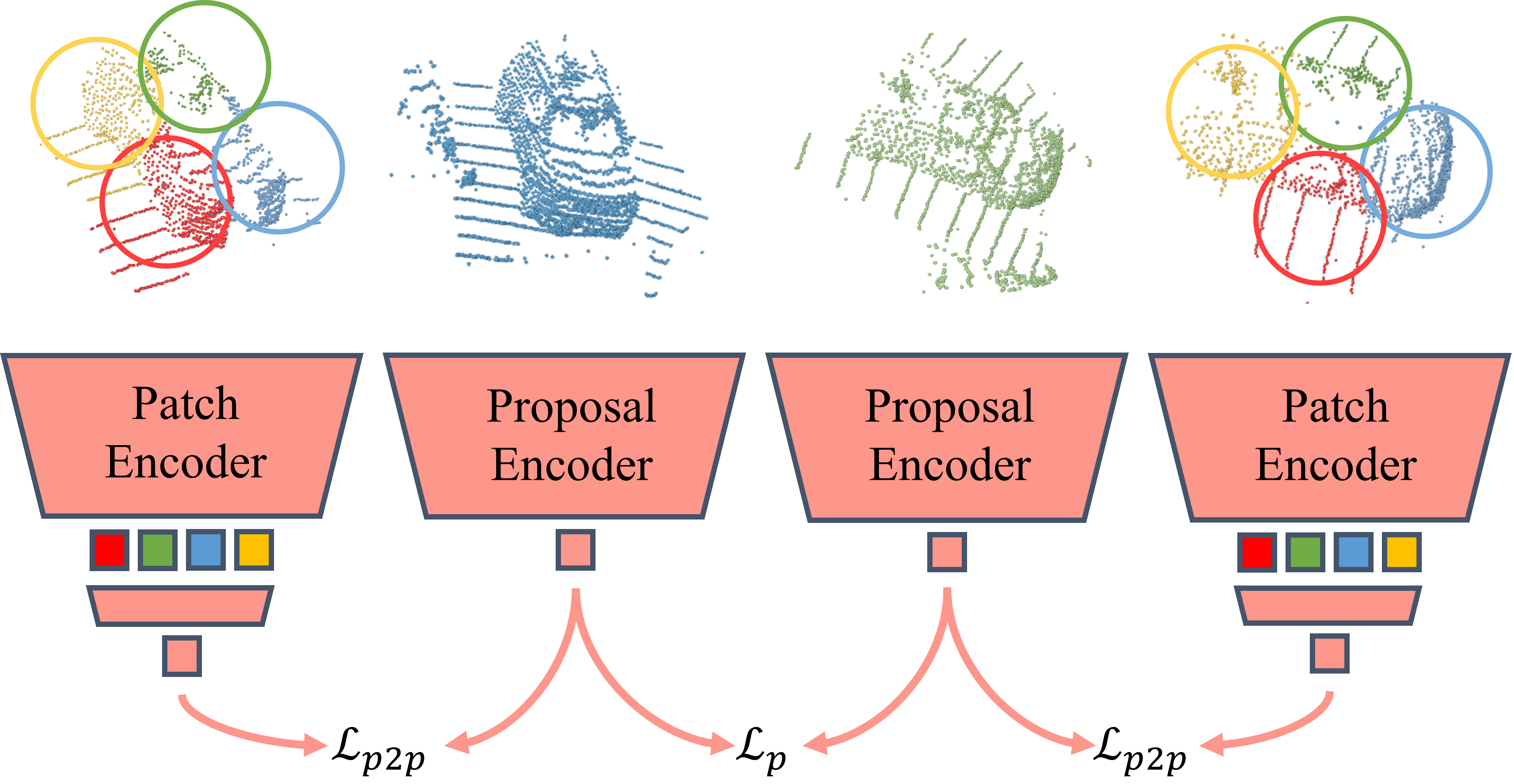}
        \\
         (a) DepthContrast & (b) ProposalContrast & (c) PatchContrast (Ours) 
    \end{tabular}
	\caption{ 
    {\bf PatchContrast overview.}
    Contrastive learning frameworks for object detection focus on different levels of abstraction. (a)~DepthContrast~\cite{depthcontrast} learns by contrasting a scene-level (global) representation.
    (b) ProposalContrast~\cite{proposalcontrast} learns by contrasting a proposal-level (object) representation.
    (c) Our \textit{PatchContrast} learns by contrasting two levels: proposals and patches which inform localization and classification respectively. 
    }
	\label{fig:overview}
\end{figure*}

We propose an additional intermediate level of abstraction, {\em patch level}, which is in-between proposals and points.
Intuitively, this level captures the connections between the components of an object, such as tires, windows, and side doors of a vehicle, which is essential for accurate classification.
By incorporating two intermediate levels of abstraction, the proposal and the patch levels, our approach overcomes the limitations of global and local approaches, boosting the discriminative power of the learned embeddings.

An additional benefit of the intermediate levels of abstraction is that they enable the division of the massive 3D scene into smaller and more manageable regions. 
This is particularly significant since despite their size, 3D scenes are sparse and contain numerous points that are not relevant to the objects of interest. 
Additionally, our approach effectively amplifies the number of negative examples available for contrastive learning, i.e., patches and proposals within a single scene, which is crucial.

We present \textit{PatchContrast}---a novel framework for contrastive learning that aims to learn accurate and efficient embedding for 3D object detection in autonomous vehicles, which realizes the proposed levels of abstraction.
As depicted in~\figref{fig:overview}, our method applies contrastive learning to proposal representations from two randomly augmented scenes and to a proposal and its composing patches representation. 
In particular, we incorporate spatial relationships among object's patches by employing an auxiliary task of masked attention, which masks a patch embedding and reconstructs it using its neighbors.

We evaluate the performance of \textit{PatchContrast} on large-scale 3D object detection datasets, including  Waymo~\cite{waymo}, KITTI~\cite{kitti1}, and ONCE~\cite{once} and demonstrate that it outperforms existing state-of-the-art self-supervised pre-training methods.
For example, \textit{PatchContrast} improves the performance of PV-RCNN~\cite{pvrcnn} by $2.40\%$, on average, over the \textit{Moderate} difficulty of KITTI.
Even when using only $50\%$ of the labeled data,  the results are improved by $1.82\%$.

Our contributions are summarized as follows:
\begin{itemize}
    \item 
    % We present a novel approach, called \textit{PatchContrast}, for self-supervised pre-training of 3D object detectors. \textit{PatchContrast} leverages available extensive  unlabeled 3D point cloud data.
    We propose \textit{PatchContrast}, a self-supervised pre-training method for 3D object detectors that leverages abundant unlabeled point cloud data.
    \item 
    % We introduce a multi-level self-supervision strategy that extracts information from proposals and patches and applies contrastive learning between them.
    % Our approach provides a viable alternative to supervision in scenarios where labeled data is limited.
    We propose a multi-level self-supervision strategy that uses proposals and patches with contrastive learning, offering an effective alternative when labeled data is scarce.
    \item 
    % We perform extensive experiments on three widely-used 3D object detection datasets: Waymo, KITTI, and ONCE. 
    % Our results demonstrate that \textit{PatchContrast} surpasses previous state-of-the-art self-supervised pre-training techniques in these benchmarks.
    We evaluate on three popular 3D object detection benchmarks—Waymo, KITTI, and ONCE—and show that \textit{PatchContrast} outperforms prior state-of-the-art self-supervised pre-training methods.
\end{itemize}

%%%%%%%%%%%%%%%%%%%%%%%%%%%%%%%%%
% section: Related Work
%%%%%%%%%%%%%%%%%%%%%%%%%%%%%%%%%
\section{Related Work}
Hereafter, we present an overview of existing research in three relevant fields.

\noindent
\textbf{Supervised 3D object detecion.}
Object Detection refers to the task of localizing and classifying objects in a given scene. 
3D methods can be divided into three categories: point-based~\cite{pointrcnn,shi2020point,yang20203dssd,yang2019std,mesika2022cloudwalker},  grid-based~\cite{yin2021center,chen2019fast,voxelrcnn,pointpillars,parta2,second,yang2018hdnet,yang2018pixor,ye2020hvnet,zheng2021cia,zheng2021se,zhou2018voxelnet, shrout2023gravos,shrout2025sfmnet}, and hybrid point-voxel approaches~\cite{pvrcnn,pvrcnn++,PVCNN}.
Point-based methods extract features directly from the raw point cloud, usually using PointNet-like architectures~\cite{qi2017pointnet,qi2017pointnet++}, while grid-based methods transform the point cloud into a regular representation, i.e., voxels or a 2D Bird-Eye View (BEV), and utilize 3D or 2D convolutional networks.
These three approaches are supervised, necessitating ground truth labels for each object in the scene during learning. 
To circumvent the laborious labeling task, self-supervised learning approaches offer an alternative. 

\noindent
\textbf{Self-supervised learning  (SSL) in 2D.}
Self-supervised learning refers to methods that involve pre-training models on large, unlabeled datasets, followed by fine-tuning on smaller labeled datasets. This approach helps in deriving valuable representations that can be effectively used for downstream learning tasks.
These techniques can be categorized as generative approaches~\cite{donahue2016adversarial,donahue2019large,mescheder2017adversarial,chen2020generative} or discriminative  methods~\cite{infoNCE,moco,mocov2,simclr,swav,grill2020bootstrap,detco}. 
For a comprehensive review, please refer to~\cite{liu2021self}.
Among discriminative methods, contrastive learning pre-training methods have demonstrated competitive performance compared to supervised pre-training~\cite{swav}.
However, the complexity of the data presents distinct challenges for 3D detection.

\begin{figure*}[t]
    \centering
	\includegraphics[width=0.95\linewidth]{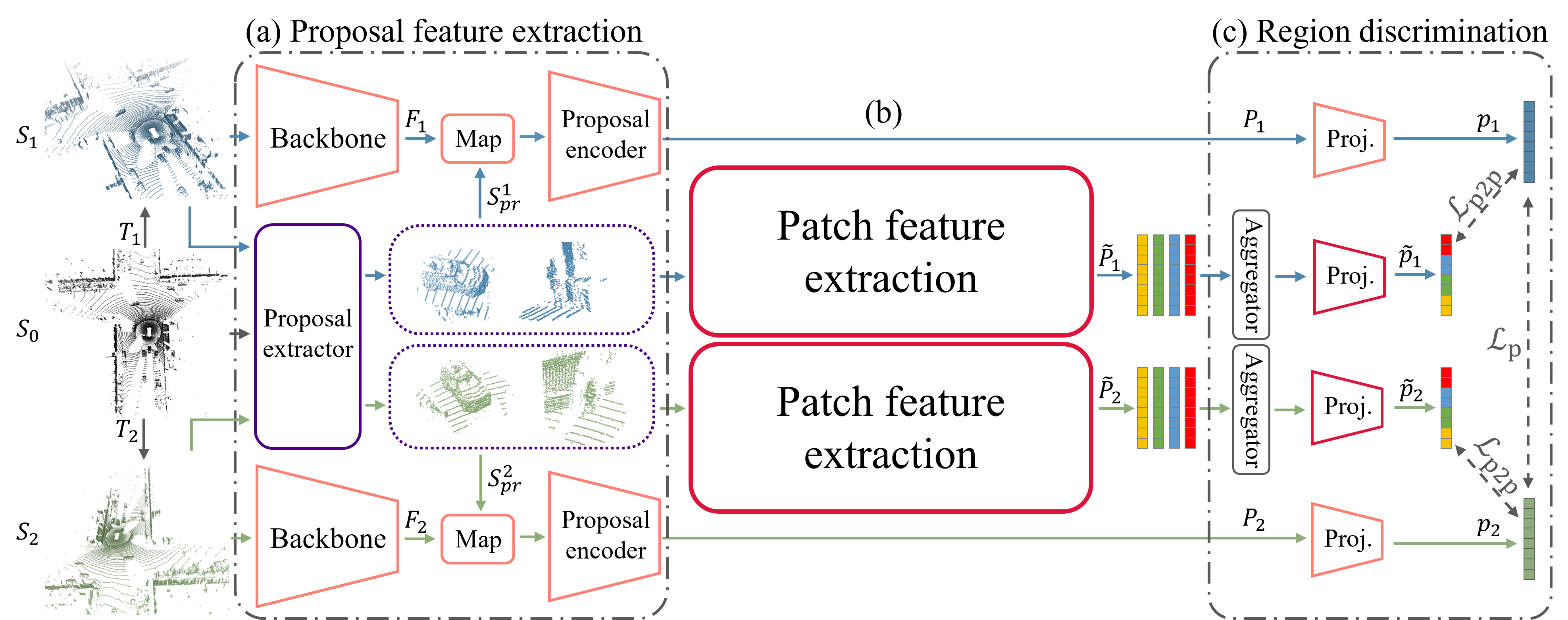} 
	\caption{ 
    {\bf PatchContrast framework.}
    (a)~Given a point cloud $S_0$ with two augmented views $S_1,S_2$, we first extract {proposals} from each view. 
    Scene view features $F_1,F_2$ are then extracted using a backbone and mapped onto the proposals $S^1_{pr},S^2_{pr}$.
    To get proposal-level features $P_1,P_2$ we feed them into a proposal encoder. 
    (b)~Simultaneously, we feed the proposals into the {patch feature extraction} module, where patches from each proposal are extracted and encoded to get patch-level features $\tilde P_1,\tilde P_2$ (see \figref{fig:patch_encoding_module}).
    (c)~Finally, the region discrimination module enforces similarity between matching proposals from the two views and between a proposal and its composing patches.
    }
	\label{fig:patchcontrast_framework}
\end{figure*}

\noindent
\textbf{Self-supervised learning  (SSL) in 3D.}
The application of contrastive learning in the context of 3D processing has received less attention compared to its use in 2D processing. 
Previous studies have mainly focused on developing representations for individual objects, which can be employed in tasks such as 
classification~\cite{STRL,hassani2019unsupervised,huang2021spatio,zhang2019unsupervised,MaskPoint,Point-MAE,IAE}, reconstruction~\cite{hassani2019unsupervised,sauder2019self,achituve2021self,wagner2023maskedfusion360}, and part segmentation~\cite{hassani2019unsupervised,alliegro2021joint,zhang2019unsupervised,sauder2019self,MaskPoint,Point-MAE}. 
Recent works have introduced techniques for 3D object detection~\cite{pointcontrast, depthcontrast, GCC-3D, ALSO}. 
DepthContrast~\cite{depthcontrast} proposes a cross-modal contrastive learning method that leverages information from both 3D point clouds and voxel representations. 
PointContrast~\cite{pointcontrast} utilizes contrastive learning on sampled points between two views of a point cloud scene. 
GCC-3D~\cite{GCC-3D} presents a two-step pre-training framework, where a 3D encoder is initially trained using a geometric-aware contrast module, and then the 3D and 2D encoders are further trained with harmonized pseudo-instance clustering. 
ALSO~\cite{ALSO} estimates the surface of a scene using an implicit representation (occupancy) and self-supervises through a reconstruction loss.
These approaches demonstrate performance improvements compared to supervised training with limited data.
However, they have not fully capitalized on the fact that the core region of interest should be at the object-level, rendering local or global representations insufficiently discriminative.
A recent work, ProposalContrast~\cite{proposalcontrast}, proposes learning an object-level representation by contrasting local subsets (proposals).
Building on this approach for localization, we show how contrasting proposals also with their constituent patches improve detection. 
Intuitively, this is so since the inter-relations between the components of an object are very informative.
We suggest a cross-level loss, encoding proposal-level information into patch-level and vice versa.

%%%%%%%%%%%%%%%%%%%%%%%%%%%%%%%%%%%% 
%%%%%% Section: Method
%%%%%%%%%%%%%%%%%%%%%%%%%%%%%%%%%%%%
\section{PatchContrast framework}
\label{sec:frame_work}
We present a novel approach for self-supervised pre-training in 3D object detection, called \textit{PatchContrast}.
It offers a promising alternative to fully-supervised methods when annotated data is scarce. 
It is built upon two main ideas: 
(1) extracting semantic features at the proposal level to aid in localization, and 
(2) extracting local features at the patch level to enable region discrimination, which is essential for classification. 
Our framework is general and can pre-train any 3D backbone used in detectors; the weights of the pre-trained backbone can then be used in detection downstream tasks.

The \textit{PatchContrast} framework has three components, shown in \figref{fig:patchcontrast_framework}: Proposal feature extraction, Patch feature extraction, and Region discrimination. 
We elaborate on each component below.

\begin{figure*}[t]
    \centering
	\includegraphics[width=.98\linewidth]{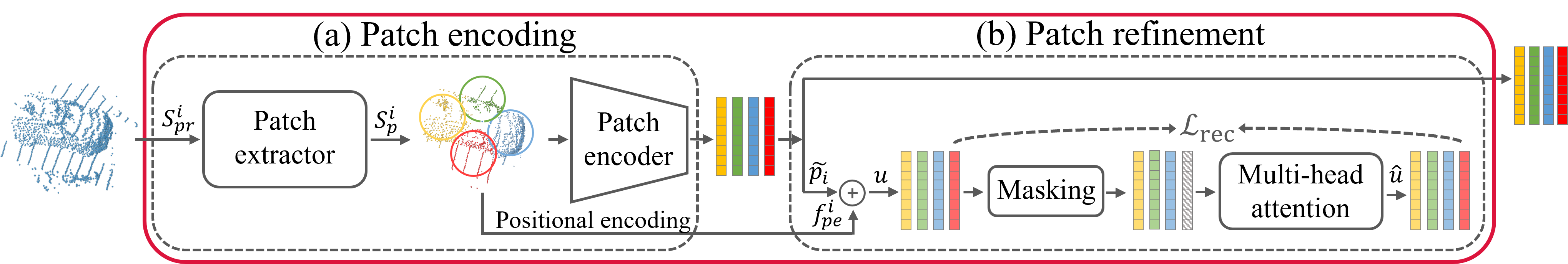}
	\caption{ 
    {\bf Patch feature extraction module.}
    (a)~Given a proposal, we extract and encode patches for patch-level representation. 
    (b)~Then, we refine the patches' representations using a masked attention auxiliary task.
    This is done by masking one patch embedding and reconstructing it by leveraging information from its neighbors' embeddings.
    }
	\label{fig:patch_encoding_module}
\end{figure*}

%%%%%%%%%%%%%%%%%%%%%%%%%%%%%%%%%%%% 
%%%%%% subSection: Prposal Extraction
%%%%%%%%%%%%%%%%%%%%%%%%%%%%%%%%%%%%
\subsection{Proposal feature extraction}
This module aims to extract small subsets, {\em proposals}, from the scene where objects are likely to appear and learn a semantically discriminative representation for each of them.  
The module takes as input three scene views:  the input cloud~$S_0$, as well as $S_1, S_2$, obtained by applying two transformations, $T_1, T_2$ on~$S_0$;
$S_i \in \mathbb{R}^{N_i \times d}$, where $N_i$ is the number of points in $S_i$ and $d$ is the number of input channels for each point (e.g., $x,y,z$ and intensity).
The transformations are sampled randomly from a family of augmentations, such as rotation, scaling, and dropout (see the supplemental for the family of augmentations). 
These different views are crucial for our contrastive learning process.

Two parallel branches are utilized: one generates proposals and the second extracts features from the scene for each proposal. 
In the first branch (middle in~\figref{fig:patchcontrast_framework}(a)), we aim to capture object-level geometric information for the two augmented views $S_1, S_2$,
by leveraging proposals.
% The proposals should capture subsets of the scene that are likely to contain objects and cover the regions of interest in the scene. 
These proposals should cover regions of interest in the scene that are likely to contain objects. 
All the proposals from both views will later be used for finding positive and negative examples in the proposal-level contrastive loss.

Specifically, the proposal extractor first employs the RANSAC algorithm~\cite{ransac} to fit a plane to the input scene $S_0$ and remove points on the background (i.e., road).
Next, it identifies the subset of non-background points in $S_0$ that are present in both views $S_1$ and $S_2$.
This subset provides a correspondence mapping between $S_1$ and $S_2$ that will be used for contrastive learning.
From this subset, $N$ query points are sampled, using Furthest Point Sampling (FPS) to encourage scene coverage.
Finally, proposal $S_{pr}^i$ is defined as a subset of points from $S_i$ that fall within a fixed-radius sphere centered around a query point.

Recall that our objective is to pre-train a 3D backbone, which encodes the scene into a spatial-aware representation vector.
Thus, the primary aim of the second branch (upper and lower in~\figref{fig:patchcontrast_framework}(a)) is to extract a local embedding for each proposal, given its respective scene view representation vector.
As current 3D detectors typically operate on a projected 2D Bird-Eye View (BEV)~\cite{yin2021center,pvrcnn,pvrcnn++,voxelrcnn}, our backbone consists of a 3D backbone and a 2D encoder. 
The weights of the 3D backbone will be transferred to the downstream detection task. 
The 3D backbone extracts features from each scene view, which are then projected onto the BEV space and encoded by the 2D encoder. The result is a feature map on a 2D grid, $F_i$, with dimension $H \times W \times C$, where $H \times W$ represents the 2D spatial subdivision of  XY plane, and $C$ represents the feature dimension. 

Finally, to learn a semantic representation for each proposal using the backbone-encoded feature vector $F_i$ and the extracted proposals from the corresponding scene view~$S_i$, we adopt an approach similar to anchor-based detectors~\cite{pvrcnn,pvrcnn++,voxelrcnn,second}. 
First, we project each proposal $S_{pr}^i$ onto the BEV space to identify relevant features in $F_i$ for each point in the proposal. 
Since $F_i$ is a grid feature map, features are assigned to each point using bilinear interpolation. 
This is done for all points in all proposals, resulting in a set of per-point proposal features. 
To obtain a semantic representation for the entire proposal, we use a point-based encoder to encode and aggregate over all the points within the proposal,
generating the proposal embedding set $P_i \in \mathbb{R}^{N \times C}$.

%%%%%%%%%%%%%%%%%%%%%%%%%%%%%%%%%%%% 
%%%%%% subSection: Patch Extraction
%%%%%%%%%%%%%%%%%%%%%%%%%%%%%%%%%%%%
\subsection{Patch feature extraction module}
\label{patches}
Given a proposal, we divide it into patches.
Each patch is expected to capture valuable local information about the individual parts of the proposals, which could aid in the classification of objects.
The two-stage process of our {\em patch feature extraction} module includes {\em patch encoding} and {\em patch refinement,} as shown in \figref{fig:patch_encoding_module}.

\noindent
\textbf{Patch encoding.}
This stage, illustrated in \figref{fig:patch_encoding_module}(a), aims to extract patches from each proposal and encode them into meaningful local representation vectors.
A patch is a subset of points of a given proposal, where the union of the patches should cover the proposal.

Our patch extractor operates in the following manner:
Initially, four keypoints are selected through the following process:
Given a spherical proposal $S_{pr}^i$ with center coordinates $q_0 = (x_0,y_0,z_0)$, we define the candidate initial centers as $(x_0 \pm t,y_0,z_0), (x_0 ,y_0 \pm t,z_0)$, where $t$ represents a translation. 
Each patch's new origin is set to be the closest point to the candidate center. 
It should be noted that due to the fact that the initial proposal's center altitude coordinate~$z_0$ is sampled from non-background points, the keypoints (and consequently, the patches) are likely to belong to an object.

To create a patch representation, we start by encoding the scene view features $F_i$ into each patch. 
This is accomplished by mapping the features from $F_i$ onto each of the patch's points, 
% $k_p$ 
feeding them through a point-based encoder, and aggregating them to generate a per-patch feature vector $\tilde P_i \in \mathbb{R}^{N \times m \times C}$, where $m$ is the number of patches.
Since the proposals are intended to capture objects, the resulting patches are highly likely to represent object components, thereby encoding valuable semantic information about them.

\noindent
\textbf{Patch refinement.}
The goal of this sub-module is to enhance the informativeness of the patch embedding (\figref{fig:patch_encoding_module}(b)). 
To achieve this objective, we suggest using an auxiliary task that modifies the representation of a patch by leveraging its neighboring patches. 
Intuitively, suppose that we mask one of the patches of a vehicle, for example, the one containing a tire. 
As humans, it is a straightforward task to determine what lies behind the mask, since the global context provides enough information to deduce spatial relationships~\cite{he2022masked}.
Our patch refinement approach is inspired by this human ability.

We propose an approach to enhance the representation of patches by incorporating spatial relationship information into their high-dimensional embeddings.
This is achieved through the combination of {\em masked attention} and positional encoding techniques.
The masked attention mechanism takes all four patch embeddings as input, masks one of them, and reconstructs it using the remaining three.
% However, when dealing with proposals containing patches from different objects
However, when proposals contain patches from different objects, the traditional masked attention approach may not yield informative results and could introduce noise.
For instance, when a proposal consists of patches from both a \textit{Vehicle} and a \textit{Pedestrian} categories, 
masking out patch embeddings from the \textit{Vehicle} and using a \textit{Pedestrian} patch for the reconstruction may not be effective.
% might not produce informative results. 
To address this issue, we add positional encoding, which encodes the normalized center coordinates of patches. 
This encoding provides spatial information about the patches' relationships, allowing the masked attention mechanism to give more weight to the correlated patches and produce more accurate representations.

Formally, let $q_p = (x_p,y_p,z_p)$ denote the center of a patch $S_p^i$ within a proposal $S_{pr}^i$ that is centered at $q_{pr} = (x_{pr},y_{pr},z_{pr})$
and let $\tilde p_i \in \tilde P_i$ be the associated encoded feature of $S_p^i$.
We obtain the normalized center of $S_p^i$ as $\bar q_p = (x_p-x_{pr}, y_p-y_{pr},z_p-z_{pr})$.
To add the relative positional encoding feature $f^i_{pe}(\bar q_p) \in \mathbb{R}^{C}$ that accounts for the spatial relationship between the patches, we project $\bar q_p$ using a single hidden layer MLP.
 As a result, the input to the masked attention is given by $u = f^i_{pe}(\bar q_p) + \tilde p_i$.

The masked attention is trained using cosine similarity as the loss function. 
Specifically, the loss encourages similarity between the original patch embedding $u$ and its reconstructed embedding $\hat{u}$ from the masked attention, defined as:
\begin{align} \label{eq:rec_loss}
    \mathcal{L}_{\text{rec}} = 1 - \frac{u^T \hat u}{\|u\| \|\hat u\|}.
\end{align}

%%%%%%%%%%%%%%%%%%%%%%%%%%%%%%%%%%%%%%%
%%%%% Subsection: Region Discrimination
%%%%%%%%%%%%%%%%%%%%%%%%%%%%%%%%%%%%%%%
\subsection{Region discrimination}
\label{subsec:losses}
The objective of this module is to learn a discriminative representation of a region, consisting of a proposal and its associated patches. 
Specifically, we are given matched proposal representation vectors from the two scene views, and their associated patches representation vectors.
We propose to employ two contrastive learning losses.
The first is between matching proposals, promoting similarity between the two views at the proposal level.
The second focuses on the relationship between a proposal in a certain view and its constituent patches, refining the representation of the proposal.

For the proposal loss, let $P_1, P_2 \in \mathbb{R}^{N \times C}$ be $N$ proposal representation sets from the scene views $S_1, S_2$ respectively. 
We project $P_1, P_2$ using a proposal projector (one hidden layer MLP) and merge the output projections to form $P$. 
Here, $P$ is a set with $2N$ samples, such that $p_{2k} \in P$ matches $p_{2k-1} \in P$ and $k \in [1,N]$ i.e.,  two consecutive elements in $P$ are positive samples, with the rest $2(N-1)$ samples as negatives.
% The rest $2(N-1)$ samples are negatives of $p_{2k}$ and $p_{2k-1}$.
% The loss between matching proposals is defined as
The proposal loss is defined as:
\begin{align} \label{eq:proposal_loss}
    \mathcal{L}_{p} = \frac{1}{2N} \sum_{k=1}^{N} \bigl( \ell(p_{2k},p_{2k-1}) + \ell(p_{2k-1},p_{2k}) \bigr).
\end{align}
In Equation~\ref{eq:proposal_loss} $\ell(p_{i},p_{j})$ is the \textit{NT-Xent} loss~\cite{simclr}, which is defined as
\begin{align}
    \ell(p_i,p_j) = -\log \frac{\exp{(\text{sim}(p_i,p_j) / \tau)}}{\sum_{k=1}^{2N} 1_{[k \neq i]} \exp{(\text{sim}(p_i,p_k) / \tau)}},
\end{align}
for a positive pair $(p_i, p_j)$. 
In this equation, $\text{sim}(p_i,p_j) = p_i^{T}p_j / \|p_i\| \|p_j\|$ is the cosine similarity between $p_i$ and $p_j$, and $\tau$ denotes the temperature parameter.

For the patch loss, let $\tilde P_1, \tilde P_2 \in \mathbb{R}^{N \times m \times C}$  be patch representations with $m$ patches in each of the $N$ proposals. 
We first aggregate all the patch representations within each proposal,  to form a single patch-level representation.
This aggregated representation can be used as a positive or negative sample against the proposals.
Next, we employ a patch projector, which is a one-hidden-layer MLP, to project the aggregated output. The output projections are then merged to obtain the final representation $\tilde P$.
Here, $\tilde P$ is a set comprising $2N$ samples, where each $\tilde p_{k} \in \tilde P$ corresponds to the matching sample for $p_k\in P$ for $k \in [1,2N]$.
Therefore, the second loss, which measures the dissimilarity between a proposal representation and its corresponding patch-level representation, is defined as:
\begin{align} \label{eq:proposal_to_patch_loss}
    \mathcal{L}_{p2p} = \frac{1}{4N} \sum_{k=1}^{2N} \bigl( \ell(p_{k},\tilde p_{k}) + \ell(\tilde p_{k},p_k) \bigr).
\end{align}

The overall loss is computed as a weighted sum of the two losses mentioned above, along with the reconstruction loss defined in \eqnref{eq:rec_loss}:
\begin{equation}
        \mathcal{L} = \lambda_1 \mathcal{L}_{p} + \lambda_2 \mathcal{L}_{p2p} + \lambda_3 \mathcal{L}_{\text{rec}}.
\label{eq:total_loss}
\end{equation} 

%%%%%%%%%%%%%%%%%%%%%%%%%%%%%%%%%%%%% 
%%%%%%%% Section: Experiments
%%%%%%%%%%%%%%%%%%%%%%%%%%%%%%%%%%%%% 
\section{Experiments}
A major advantage of self-supervised pre-training is the ability to transfer knowledge gained from large unlabeled datasets to small annotated ones.
To evaluate our method, we conducted experiments on the most widely used benchmarks for 3D object detection in autonomous driving, which include Waymo~\cite{waymo}, KITTI~\cite{kitti1}, and ONCE~\cite{once}.
For all the experiments, we adopt Waymo for the pre-training, where we evaluate our pre-trained backbone generalizability in both in-domain and out-of-domain by transfer learning to KITTI and ONCE.
Specifically, we fine-tune several detectors~\cite{yin2021center,pvrcnn,second} on different detection benchmarks and show that our approach outperforms SoTA approaches.

Hereafter, we present the datasets and metrics, implementation details, results, and ablation study. 
Additional results are provided in the supplemental materials.

\noindent
\textbf{Datasets and metrics.}
\textit{Waymo}~\cite{waymo} contains $158,081$, and $40,077$ LiDAR samples for training and validation.
Average Precision (AP) and Average Precision weighted by Heading (APH) are used for evaluation. 

\textit{KITTI}~\cite{kitti1} contains $3,712$, and $3,769$ examples for training and validation~\cite{chen20153d}. 
A mean Average Precision (mAP) with $40$ recall positions is used for evaluation~\cite{simonelli2019disentangling}.

\textit{ONCE}~\cite{once} contains $4,961$, and $3,321$ scenes for supervised training.
For unsupervised pre-training, the dataset contains $3$ subsets: $U_{s}$, $U_{m}$, and $U_{l}$, corresponding to $100K$, $500K$ and $1M$ scenes, respectively.
An orientation-aware AP is used for evaluation.

\subsection{Implementation Details}
\textbf{Pre-training details.} 
We pre-train two standard voxel-based sparse convolution backbones: {\em VoxelBackBone8x}~\cite{pvrcnn,pvrcnn++,voxelrcnn,second} and {\em VoxelResBackBone8x}~\cite{yin2021center,zhou2018voxelnet}.
For all the downstream tasks, we pre-train on the Waymo train set ($\sim 158K$ frames) for $30$ epochs, with no annotations.
We set for each frame $d=4$ input channels, i.e., $x,y,z,intensity$.
Even though Waymo has another feature (\textit{elongation}), we chose to skip it since other datasets do not have it.
The standard 2D encoder~\cite{yin2021center,pvrcnn,voxelrcnn,second} is utilized for the BEV features, enforcing $C=512$ for the proposals and patches.
We sample $N=1024$ proposals with radius $R_{pr}=1$ and set the maximal number of points in each proposal to be $k_{pr}=16$.
The Transformer in \cite{proposalcontrast} is used as the \textit{Proposal Encoder}.
The patches' initial centers are translated from the proposal's origin with $t=1/3$ and have a radius of $R_p=1/3$.
A maximum of $k_p=8$ points are sampled for each patch.
For the \textit{Patch Encoder}, PointNet~\cite{qi2017pointnet} is utilized.
The representation vectors $p,\tilde p$ of the proposals and patches in Equations \ref{eq:proposal_loss} and \ref{eq:proposal_to_patch_loss} are of dimension $128$ with temperature~$\tau=0.1$.
For the loss weights in Equation~\ref{eq:total_loss} we set $\lambda_1 = 1$, $\lambda_2 = 1$, and $\lambda_3 = 0.05$.
For training, we use Adam optimizer, with a cosine learning rate scheduler as in \cite{proposalcontrast,pvrcnn}, and set the maximum learning rate to 0.003.
We train the model with a batch size of $16$ across $8 \times$ A100.
% NVIDIA A100-SXM4-80GB GPUs.

\noindent
\textbf{Data augmentations.} 
% Data augmentation plays a major role in SSL settings since it sets the variations in the data that the model will be exposed to.
We apply standard data augmentations, such as random flipping with a probability of $0.5$, random scaling drawn from $[0.95, 1.05]$, and random rotation around the $z$ axes, with angle drawn from $[-45^{\circ}, 45^{\circ}]$ and with $0.5$ probability.
We randomly drop up to $20\%$ of the points from the scene and insert random noise into the remaining points.
Specifically, we add Gaussian noise $n \sim {\mathcal{N}(0,\sigma^2)}$ with $\sigma$ drawn from $[0, 0.015]$ for the coordinates and from $[0, 0.01]$ for the intensity.
We also apply a random cuboid~\cite{depthcontrast} with a minimum area of $0.75$ and randomly drop scene patches of a maximum $20\%$ of the scene.

%%%%%%%%%%%%%%%%%%%%%%%%%%%%%%%%%%%
%%%% subsection: Label efficiency for 3D detection
%%%%%%%%%%%%%%%%%%%%%%%%%%%%%%%%%%%
\subsection{In-domain 3D detection results (on Waymo)}
We evaluate the performance of our pre-trained backbones under a label-efficient setting i.e., with different amounts of labeled data.
Specifically, we split the Waymo training set into two groups of $399$ sequences, which are equal to about $80K$ frames.
The first $399$ sequences are used for pre-training, and different amounts of labels from the remaining $399$ sequences are used for finetuning.
We finetune PV-RCNN, and SECOND with $1\%$ ($0.8K$ frames), $5\%$ ($4K$ frames), and $10\%$ ($8K$ frames).
We adopt the 1x scheduler ($12$ epochs) for finetuning and we evaluate over the validation set.
\tabref{tab:waymo_label_efficiency} reports the AP and APH of the Level-2 difficulty, averaged over $3$ trials.
The results show that our method provides significant benefits when fewer labeled samples are available.
More results are in the supplemental.

Furthermore, we adhere to the standard protocol in~\cite{openpcdet2020}, which involves an initial pre-training phase on the entire Waymo training dataset, followed by fine-tuning with $20\%$ of labeled examples ($\sim 31.6K$ scenes) from the same training set for $30$ epochs. 
The evaluation is then performed on the validation set.
\tabref{tab:finetune_waymo_results} reports the results on Level-2, compared to other SoTA pre-training methods, namely GCC-3D~\cite{GCC-3D} and ProposalContrast~\cite{proposalcontrast}, using two different detectors: PV-RCNN and CenterPoint.
The results show that our approach achieves improved performance over the baseline (training from scratch), and outperforms other methods. 
We note that we pre-trained solely on the training split, whereas other methods also used the validation set.
See supplemental for results on Level-1.

\begin{table}[t]
    \scriptsize
    \setlength\tabcolsep{2pt} % default value: 6pt
    \centering
    \begin{tabular}{c | c | c c | c c | c c | c c}
         \toprule
            \multirow{2}{*}{Labels} & \multirow{2}{*}{Method} & \multicolumn{2}{c|}{Overall} & \multicolumn{2}{c|}{{\em Vehicle}} & \multicolumn{2}{c|}{{\em Pedestrian}} & \multicolumn{2}{c}{{\em Cyclist}} \\
             &  & \multicolumn{1}{c}{mAP} & \multicolumn{1}{c|}{mAPH} & \multicolumn{1}{c}{AP} & \multicolumn{1}{c|}{APH} & \multicolumn{1}{c}{AP} & \multicolumn{1}{c|}{APH} & \multicolumn{1}{c}{AP} & \multicolumn{1}{c}{APH}\\  
        \midrule
            \multirow{7}{*}{1\%} & CenterPoint & 27.17 & 23.75 & 27.60 & 27.03 & 32.29 & 23.78 & 21.64 & 20.43 \\
             & Ours & \textbf{34.43} & \textbf{30.29} & 37.55 & 36.91 & 38.03 & 28.07 & 27.71 & 25.88 \\
            \cmidrule{2-10}
             & PV-RCNN & 26.87 & 18.38 & 41.96 & 34.85 & 23.72 & 11.95 & 14.92 & 8.35 \\
              & Ours & \textbf{38.40} & \textbf{25.18} & 48.70 & 39.46 & 34.91 & 17.06 & 31.59 & 19.03 \\
            \cmidrule{2-10}
             & SECOND & 20.31 & 14.83 & 33.93 & 30.85 & 24.31 & 12.34 & 2.67 & 1.31 \\
             & Ours & \textbf{30.05} & \textbf{21.42} & 39.67 & 36.55 & 30.23 & 15.65 & 20.25 & 12.05 \\
        \midrule
            \multirow{7}{*}{5\%} & CenterPoint & 49.74 & 46.90 & 49.83 & 49.21 & 48.00 & 41.40 & 51.39 & 50.09 \\
             & Ours & \textbf{51.40} & \textbf{48.48} & 51.79 & 51.17 & 48.12 & 41.33 & 54.30 & 52.94 \\
            \cmidrule{2-10}
             & PV-RCNN & 51.97 & 36.27 & 58.99 & 57.85 & 48.62 & 24.33 & 48.28 & 26.64 \\
              & Ours & \textbf{54.66} & \textbf{38.21} & 60.20 & 59.28 & 50.44 & 25.68 & 53.34 & 29.68 \\
            \cmidrule{2-10}
             & SECOND & 40.49 & 28.65 & 48.44 & 47.45 & 39.96 & 20.44 & 33.08 & 18.04 \\
             & Ours & \textbf{45.00} & \textbf{30.90} & 51.34 & 50.44 & 42.79 & 21.08 & 40.89 & 21.19 \\
        \midrule
            \multirow{7}{*}{10\%} & CenterPoint & 55.74 & 53.00 & 54.54 & 53.95 & 54.12 & 47.90 & 58.56 & 57.14 \\
             & Ours & \textbf{56.24} & \textbf{53.41} & 55.48 & 54.89 & 54.29 & 47.75 & 58.94 & 57.59 \\
            \cmidrule{2-10}
            & PV-RCNN & 56.38 & 38.77 & 61.77 & 60.94 & 52.74 & 26.2 & 54.65 & 29.17 \\
              & Ours & \textbf{57.65} & \textbf{39.43} & 62.29 & 61.56 & 53.85 & 26.69 & 56.79 & 30.06 \\
            \cmidrule{2-10}
             & SECOND & 46.12 & 32.56 & 52.97 & 52.14 & 44.98 & 23.22 & 40.40 & 22.32 \\
             & Ours & \textbf{48.89} & \textbf{34.17} & 54.75 & 53.99 & 46.86 & 23.50 & 45.06 & 25.03 \\
         \bottomrule
    \end{tabular}
    \caption{
    \textbf{Data efficiency on Waymo.} 
    The less labeled data available, the more beneficial our approach becomes. 
    Thus, our approach provides a valuable alternative to supervision when labeled data is scarce.
    }
    \label{tab:waymo_label_efficiency}
\end{table}

\begin{table}[t]
    \scriptsize
    \setlength\tabcolsep{3pt} % default value: 6pt
    \centering
    \begin{tabular}{l | c c | c c | c c | c c}
        \toprule
            \multirow{2}{*}{Method} & \multicolumn{2}{c|}{Overall} & \multicolumn{2}{c|}{{\em Vehicle}} & \multicolumn{2}{c|}{{\em Pedestrian}} & \multicolumn{2}{c}{{\em Cyclist}} \\
             & \multicolumn{1}{c}{mAP} & \multicolumn{1}{c|}{mAPH} & \multicolumn{1}{c}{AP} & \multicolumn{1}{c|}{APH} & \multicolumn{1}{c}{AP} & \multicolumn{1}{c|}{APH} & \multicolumn{1}{c}{AP} & \multicolumn{1}{c}{APH}\\  
        \midrule
            PV-RCNN~$\dag$ & 59.84 & 56.23  & 64.99 & 64.38  & 53.80 & 45.14  & 60.72 & 59.18 \\
            PV-RCNN & 64.84 & 60.86 & 67.44 & 66.80 & 63.70 & 53.95 & 63.39 & 61.82 \\
            GCC-3D & 61.30 & 58.18  & 65.65 & 65.10  & 55.54 & 48.02  & 62.72 & 61.43 \\
            ProposalContrast & 62.62 & 59.28  & 66.04 & 65.47  & 57.58 & 49.51  & 64.23 & 62.86  \\
            Ours & \textbf{67.91} & \textbf{64.14} & 68.40 & 67.84 & 66.62 & 57.48 & 68.72 & 67.11 \\
        \midrule
            CenterPoint~$\dag$ & 63.46 & 60.95 & 61.81 & 61.30 & 63.62 & 57.79 & 64.96 & 63.77 \\
            CenterPoint & 66.48 & 64.01 & 64.91 & 64.42 & 66.03 & 60.34 & 68.49 & 67.28 \\
            GCC-3D & 65.29 & 62.79 & 63.97 & 63.47 & 64.23 & 58.47 & 67.68 & 66.44 \\
            ProposalContrast & 66.42 & 63.85 & 64.94 & 64.42 & 66.13 & 60.11 & 68.19 & 67.01  \\
            Ours & \textbf{67.02} & \textbf{64.57} & 64.73 & 64.25 & 67.10 & 61.45 & 69.22 & 68.01 \\
         \bottomrule
    \end{tabular}
    \caption{
    \textbf{3D object detection on $20\%$ Waymo using OpenPCDet's protocol.} 
    Our approach outperforms previous methods.
    $\dag$ reported by~\cite{GCC-3D}.
    }
    \label{tab:finetune_waymo_results}
\end{table}

\begin{table}
    \scriptsize
    \setlength\tabcolsep{6pt} % default value: 6pt
    \centering
    \begin{tabular}{ c | c | c | c c c }
         \toprule
            Labels & Method & mAP & {\em Car} & {\em Pedestrian} & {\em Cyclist} \\
        \midrule
            \multirow{3}{*}{20\%} & PV-RCNN & 66.71 & 82.52 & 53.33 & 64.28 \\
            & ProposalContrast & 68.13  & 82.65 & 55.05 &  66.68 \\
            & PatchContrast (Ours) & \textbf{70.75} & 82.63  & 57.77 &  71.84 \\ % Ours (ckp10, best)
        \midrule
            \multirow{3}{*}{50\%} & PV-RCNN & 69.63 & 82.68  & 57.10  & 69.12 \\
            & ProposalContrast & 71.76 & 82.92 &  59.92  & 72.45 \\
            & PatchContrast (Ours) & \textbf{72.39} & 84.47  & 60.76 &  71.94 \\ % Ours (ckp15, best)
        \midrule
            \multirow{7}{*}{100\%} & PV-RCNN & 70.57  & 84.50 & 57.06 & 70.14 \\
            & GCC-3D & 71.26 & - & - & -  \\
            & STRL & 71.46  & 84.70 & 57.80 & 71.88  \\
            & PointContrast & 71.55  & 84.18 & 57.74 &  72.72  \\
            & ProposalContrast & 72.92  & 84.72 &  60.36 &  73.69 \\
            & ALSO & 72.96  & 84.68  & 60.16  & 74.04  \\
            & PatchContrast (Ours) & \textbf{72.97} & 84.67 & 59.92 & 74.33 \\ %  Ours (best epoch)
         \bottomrule
    \end{tabular}
    \caption{
    \textbf{Transfer learning on KITTI (Moderate).} 
   Our approach outperforms the alternatives on the moderately difficult KITTI validation set, especially when a limited number of labels are available for training.
   }
    \label{tab:finetune_kitti_results}
\end{table}

%%%%%%%%%%%%%%%%%%%%%%%%%%%%%%%%%%%
%%%% subsection: Transfer learning
%%%%%%%%%%%%%%%%%%%%%%%%%%%%%%%%%%%
\subsection{Transfer learning for 3D detection (out of domain)}
In this experiment, we first pre-train on the Waymo training set and then fine-tune on the specific dataset we evaluate.

\noindent
\textbf{Transfer learning on KITTI dataset.}
We follow the setup proposed in~\cite{proposalcontrast}: 
After pre-training on Waymo training split, we finetune with different amounts of labeled data of KITTI's train split and report on the entire validation split.
Specifically, the train set is split into $20\%$, $50\%$, and $100\%$, resulting in $0.7K$, $1.9K$ and $3.7K$ scenes, respectively.
\tabref{tab:finetune_kitti_results} reports the results on the KITTI 3D detection benchmark for PV-RCNN detector.
It shows that for all the splits our approach improves the results.

An important observation is that the improvement is higher whenever less labeled data is available.
When all training examples are available ($100\%$) we achieve on-par results with \cite{proposalcontrast}.
However, with a limited number of annotated samples, we achieve better results than training from scratch on the full annotated dataset.
Specifically, on the \textit{Moderate} level, we improve the baseline (with $100\%$ annotations) by $0.18$ and $1.82$ when trained with only $20\%$ and $50\%$ annotations, respectively.
See the supplementary materials for similar results of other difficulty levels.

% \vspace{0.2in}
\begin{table}[t]
    \scriptsize
    \setlength\tabcolsep{4pt} % default value: 6pt
    \centering
    \begin{tabular}{l | c | c | c c c }
        \toprule
            \multirow{2}{*}{Method} & Pre-trained & \multirow{2}{*}{mAP} & \multirow{2}{*}{{\em Vehicle}} & \multirow{2}{*}{{\em Pedestrian}} & \multirow{2}{*}{{\em Cyclist}} \\
             & dataset & & & & \\
        \midrule
            SECOND & - & 51.89 & 71.19 & 26.44 & 58.04 \\ 
            BYOL & $U_{s}$ & 46.04 & 68.02 & 19.50 & 50.61 \\
            PointContrast & $U_{s}$ & 49.98 & 71.07 & 22.52 & 56.36 \\
            SwAV & $U_{s}$ & 51.96 & 72.71 & 25.13 & 58.05 \\
            DeepCluster & $U_{s}$ & 52.06 & 73.19 & 24.00 & 58.99 \\
            ALSO & $U_{s}$ & 52.68 & 71.73 & 28.16 & 58.13 \\
            DepthContrast & Waymo & 52.21 & 71.93 & 26.77 & 57.93 \\
            PatchContrast (Ours) & Waymo & \textbf{55.15} & 72.95 & 32.68 & 59.83 \\ 
        \bottomrule
    \end{tabular}
    \caption{
    \textbf{Transfer learning on ONCE.}
    Our framework outperforms previous approaches, even when pre-trained on a different dataset.
    }
    \label{tab:once_results}
\end{table}

\label{subsec:once_results}
\noindent
\textbf{Transfer learning on ONCE dataset.}
We further evaluate our backbones' generalizability on the  ONCE dataset.
We pre-train on Waymo, as in the previous experiment, finetune SECOND on ONCE's train set, and evaluate on the validation set. 
\tabref{tab:once_results} reports on the official self-supervised benchmark results of~\cite{once} and compares our results to those of the reported SoTA approaches.
It clearly shows that even when pre-trained on a different dataset, our method improves the baselines and outperforms previous methods.

%%%%%%%%%%%%%%%%%%%%%%%%%%%%%%%%%%%
%%%% subsection: Additional evaluation
%%%%%%%%%%%%%%%%%%%%%%%%%%%%%%%%%%%
\subsection{Additional evaluation}
In addition to the standard evaluation, we propose two methods to assess the quality of the learned embedding before fine-tuning, which may overwrite the weights.

\begin{figure}[t]
    \centering
	\includegraphics[width=\linewidth]{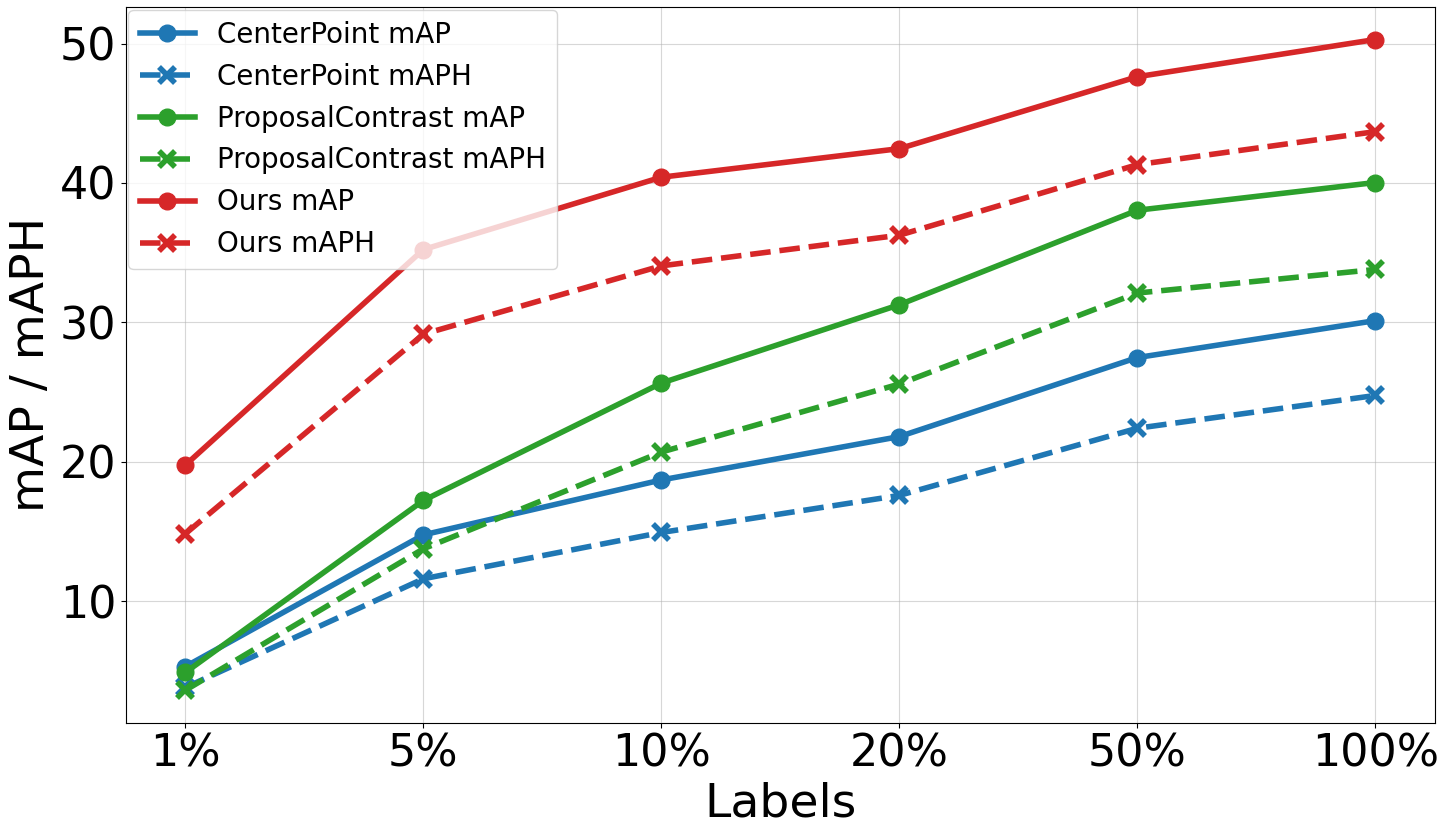}
	\caption{
    \textbf{Data efficiency on Waymo with frozen features.} 
    When using our learned features and training the detection head on only $10\%$ of the data, our approach already outperforms the SoTA trained on $100\%$.
    % Our backbone's learned embedding encapsulates meaningful information about the scenes without fine-tuning.
    }
	\label{fig:frozen_backbone_results}
\end{figure}

\begin{figure}[t]
    \centering
	\includegraphics[width=\linewidth]{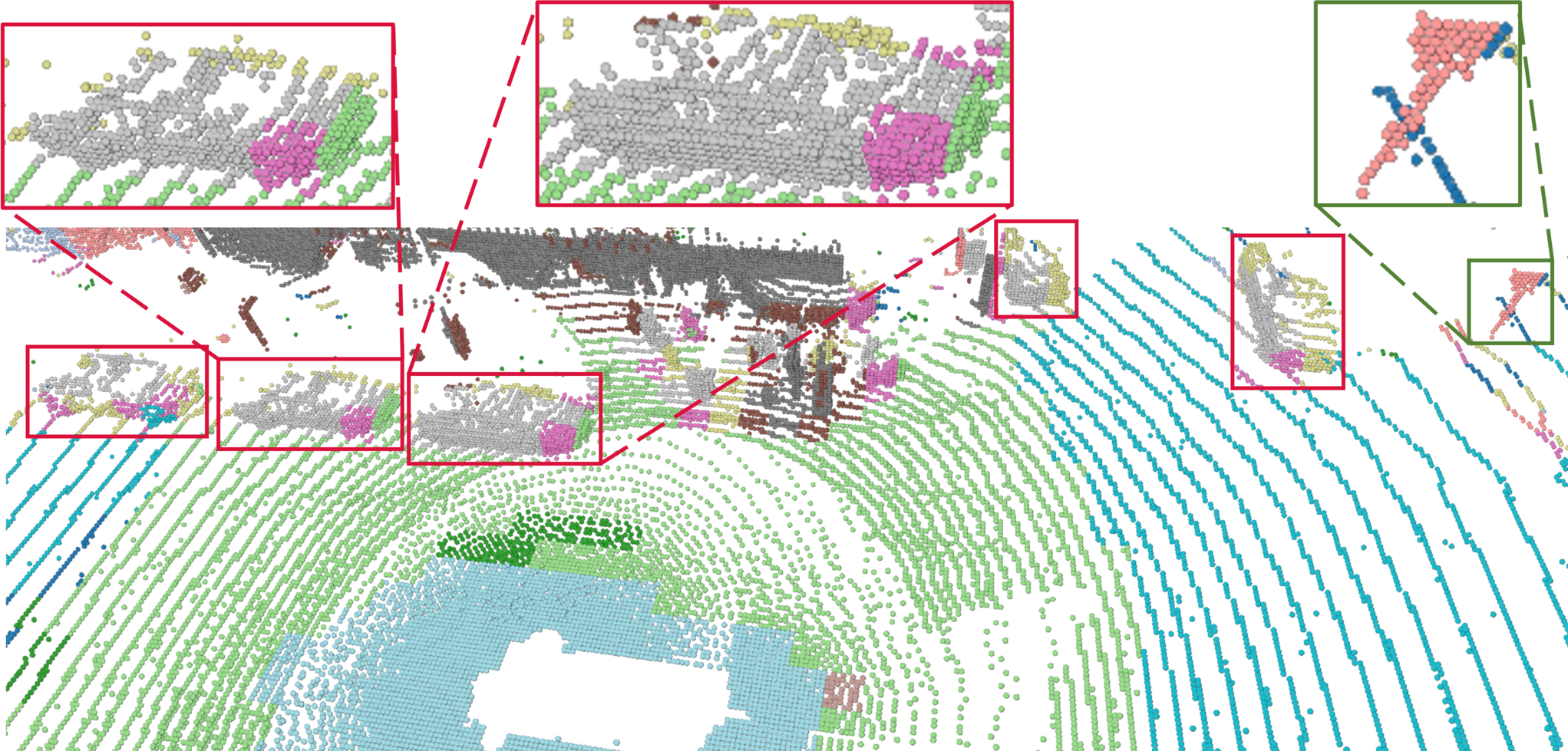}
	\caption{
    \textbf{Qualitative evaluation.}
    Clustering the embeddings reveals semantically meaningful object components, such as cars and signs--discovered without any supervision.
    % When clustering the embeddings, we can observe semantically meaningful object components, such as cars and signs. Remarkably, these components are discovered without any supervision. 
    }
	\label{fig:qualitative_results}
\end{figure}

\noindent
\textbf{Quantitative evaluation.}
It has been shown that as the amount of labeled data and the number of fine-tuning iterations increase, the impact of pre-training diminishes~\cite{he2019rethinking}.
To evaluate the quality of the learned embeddings before fine-tuning modifies the backbone's weights, we adopt a protocol similar to the linear classification approach in~\cite{simclr,moco}: we freeze the backbone features and train a detection head on top of them. 
Specifically, we perform unsupervised pre-training on half of the Waymo training split, freeze the extracted features, train the CenterPoint detector on the remaining half, and evaluate performance on the validation set.
\figref{fig:frozen_backbone_results} presents results on the Level-2 difficulty of Waymo for different percentages of labeled data, with each experiment averaged over three trials (using the 1x scheduler). 
The results show that our learned embeddings capture meaningful information about point cloud scenes even without fine-tuning.
Moreover, even when using our features and training the detection head on only $10\%$ of the data, our approach already outperforms the SoTA trained on $100\%$.
The full results are provided in the supplemental material.

\noindent
\textbf{Qualitative evaluation.}
We clustered the scene's points in the embedding space using the k-means algorithm.
\figref{fig:qualitative_results} illustrates a typical outcome, obtained from our pre-trained backbone. 
The results demonstrate that the embedding space of the backbone captures a semantic representation of objects.
For instance, cars are clustered together (grey, golden, and pink). 
Moreover, our approach captures local information about objects and their constituent parts. 
As an example, the front right wheels of all cars are grouped in the same cluster, indicated by pink. 
This ability to capture the structural information can be attributed to our patch-level abstraction learning.
Furthermore, since our approach does not rely on labeled data, it can locate objects that do not fall into pre-defined classes.
For instance, it detects the traffic sign within the green rectangle on the right, even though it is not part of the pre-defined categories.

%%%%%%%%%%%%%%%%%%%%%%%%%%%%%%%% 
% Ablation studies
%%%%%%%%%%%%%%%%%%%%%%%%%%%%%%%%
\subsection{Ablation studies}
We demonstrate the importance of our Patch refinement module along with other parameter choices.
We use CenterPoint, pre-training on the Waymo train split, fine-tuning on $20\%$ of the Waymo train split for $30$ epochs, and evaluating on the full validation set.
\tabref{sup:ablation_refinement} highlights the benefits of our Patch refinement module. 
By incorporating our proposed auxiliary task to refine patch embeddings, we learn a more discriminative representation. 
This improvement is particularly evident in the challenging \textit{Pedestrian} class.

\begin{table}[t]
    \scriptsize
    \setlength\tabcolsep{4pt} % default value: 6pt
    \centering
    \begin{tabular}{c | c c | c c | c c | c c}
        \toprule
            Patch & \multicolumn{2}{c|}{Overall} & \multicolumn{2}{c|}{{\em Vehicle}} & \multicolumn{2}{c|}{{\em Pedestrian}} & \multicolumn{2}{c}{{\em Cyclist}} \\
            Refinement & \multicolumn{1}{c}{mAP} & \multicolumn{1}{c|}{mAPH} & \multicolumn{1}{c}{AP} & \multicolumn{1}{c|}{APH} & \multicolumn{1}{c}{AP} & \multicolumn{1}{c|}{APH} & \multicolumn{1}{c}{AP} & \multicolumn{1}{c}{APH} \\  
        \midrule
            \xmark & 66.22 & 63.80 & 64.38 & 63.89 & 65.69 & 60.05  & 68.60 & 67.46 \\
            \cmark & \textbf{67.02} & \textbf{64.57} & 64.73 & 64.25 & 67.10 & 61.45 & 69.22 & 68.01 \\
        \bottomrule
    \end{tabular}
    \caption{
    \textbf{Patch refinement importance.} Our auxiliary task is beneficial for improving the average detection score, especially for the challenging \textit{Pedestrian} class.
    }
    \label{sup:ablation_refinement}
\end{table}

We further evaluate the number of patches extracted from each proposal.
We change the patches' radius, $R_p$, and the translation scalar, $t$, accordingly to get the same coverage of the proposal without increasing the overlap between the patches.
We decrease the number of proposals, $N$, by the same ratio of the increase in the number of patches, due to memory limitation.
\tabref{sup:ablation_modules} shows that $m=4$ patches with $N=1024$ proposals provide the best results.

\begin{table}[t]
    \scriptsize
    \setlength\tabcolsep{2pt} % default value: 6pt
    \centering
    \begin{tabular}{ c | c | c | c | c c | c c | c c | c c }
        \toprule
            \multirow{2}{*}{$m$} & \multirow{2}{*}{$R_p$} & \multirow{2}{*}{$t$} & \multirow{2}{*}{$N$} & \multicolumn{2}{c|}{Overall} & \multicolumn{2}{c|}{{\em Vehicle}} & \multicolumn{2}{c|}{{\em Pedestrian}} & \multicolumn{2}{c}{{\em Cyclist}} \\
            & & & & \multicolumn{1}{c}{mAP} & \multicolumn{1}{c|}{mAPH} & \multicolumn{1}{c}{AP} & \multicolumn{1}{c|}{APH} & \multicolumn{1}{c}{AP} & \multicolumn{1}{c|}{APH} & \multicolumn{1}{c}{AP} & \multicolumn{1}{c}{APH} \\ 
        \midrule
            0 & 0.000 & 0.000 & 1024 & 66.22 & 63.80 & 64.38 & 63.89 & 65.69 & 60.05  & 68.60 & 67.46 \\
            6 & 0.272 & 0.385 & 682 & 66.34 & 63.94  & 64.24 & 63.75 & 66.21 & 60.64 & 68.58 & 67.42 \\
            4 & 0.333 & 0.333 & 1024 & \textbf{67.02} & \textbf{64.57} & 64.73 & 64.25 & 67.10 & 61.45 & 69.22 & 68.01 \\
        \bottomrule
    \end{tabular}
    \caption{
    \textbf{Different number of patches and proposals.}
    Extracting $4$ patches from $1024$ proposals provides the best results.
    }
    \label{sup:ablation_modules}
\end{table}

%%%%%%%%%%%%%%%%%%%%%%%%%%%%%%%% 
% Conclusion
%%%%%%%%%%%%%%%%%%%%%%%%%%%%%%%%
\section{Conclusion}
We introduced a self-supervised pre-training framework, called \textit{PatchContrast}, for 3D object detection.
\textit{PatchContrast} incorporates two levels of abstraction: proposal level and patch level, enabling the learning of discriminative embedding.
Patches reveal the interrelation between the components of the object within the proposal.
These inter-relations, refined through an auxiliary task, allow for contrasting different representations of the proposal, considering both its components and the proposal as a whole. 
% Through the combination of the levels of abstraction, \textit{PatchContrast} effectively learns discriminative representations for the detection task.

We validated the efficacy of \textit{PatchContrast}  on three widely-used 3D object detection datasets, surpassing previous approaches.
Additionally, our results demonstrated that pre-training on large unlabeled data can enhance detection accuracy, particularly when labeled data is limited.

%%%%%%%%%%%%%%%%%%%%%%%%%%%% 
%%%%%%% Acknowledgement
%%%%%%%%%%%%%%%%%%%%%%%%%%%%
\clearpage
\noindent
{\bf Acknowledgement.}
This research was partially supported by the Israeli Smart Transportation Research Center (ISTRC), the Advanced Defense Research Institute (ADRI) at the Technion, the Israel Science Foundation (grant No. 2329/22), and the European Union’s Horizon 2020 research and innovation programme under the Marie Skłodowska-Curie grant agreement No. 893465.

%%%%%%%%%%%%%%%%%%%%%%%%%%%% 
%%%%%%% References
%%%%%%%%%%%%%%%%%%%%%%%%%%%%
{
    \small
    \bibliographystyle{ieeenat_fullname}
    \bibliography{references}
}

%%%%%%%%%%%%%%%%%%%%%%%%%%%% 
%%%%%%% Supplemental (comment before submitting)
%%%%%%%%%%%%%%%%%%%%%%%%%%%%
\clearpage
\section{Supplemental Materials}
We provide additional qualitative and quantitative results.

\subsection{Qualitative results}
\label{sup:qualitative_results}
\begin{figure}[t]
    \centering
	\includegraphics[width=.96\linewidth]{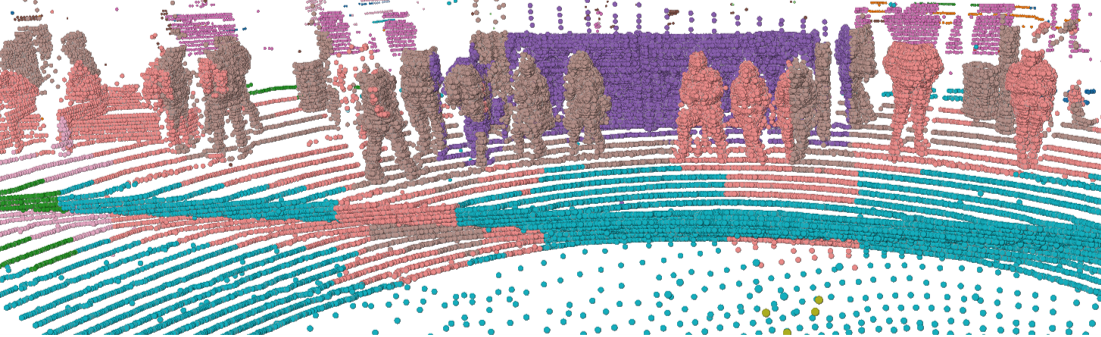}
	\caption{
    \textbf{Qualitative result.}
    A scene with pedestrians.
    }
	\label{fig:pedestrians}
\end{figure}

\begin{figure}[t]
    \centering
	\includegraphics[width=.96\linewidth]{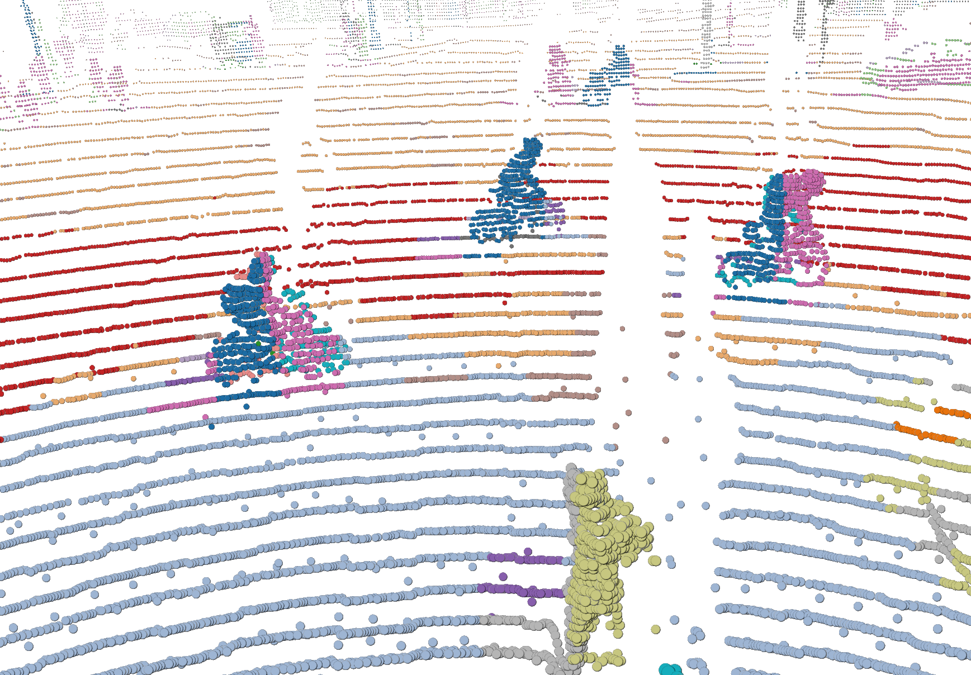}
	\caption{
    \textbf{Qualitative result.}
    A scene with cyclists.
    }
	\label{fig:cyclists}
\end{figure}

\begin{figure}[th]
    \centering
	\includegraphics[width=.96\linewidth]{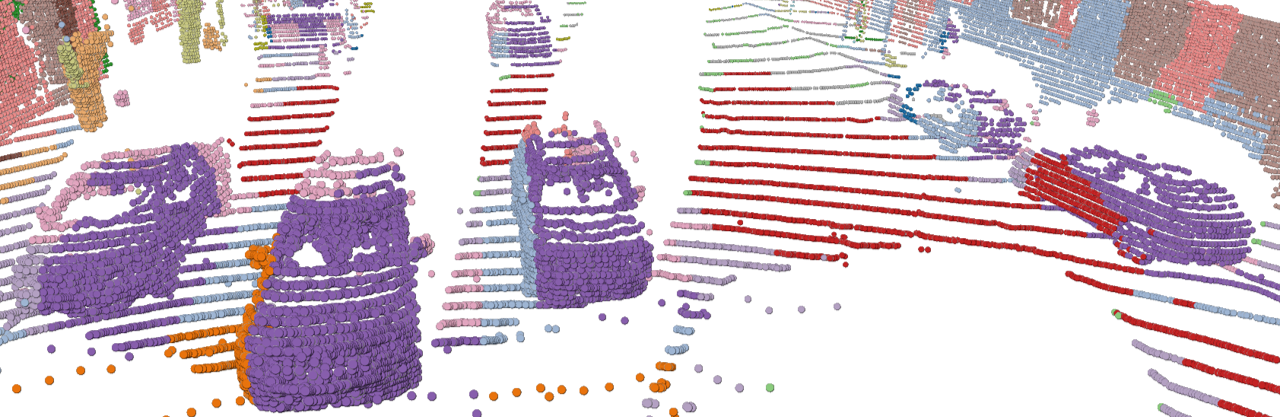}
	\caption{
    \textbf{Qualitative result.}
    A scene with vehicles.
    }
	\label{fig:cars}
\end{figure}

To qualitatively evaluate our self-supervised pre-training backbone, we used a K-means algorithm on the backbone's embedding right after the pre-training stage (without finetuning). 
In particular, we extracted the 2D-grid BEV embedding of each scene from $1\%$ samples of Waymo~\cite{waymo} using the pre-trained backbone. 
We extracted each pixel embedding from the BEV of each scene to form a dataset. 
We used the K-means algorithm with $20$ clusters to cluster these embeddings and project each BEV pixel's resulting cluster back onto their corresponding locations in the original point cloud.
\figref{fig:pedestrians}, \figref{fig:cyclists}, and \figref{fig:cars} depict qualitative examples for Pedestrians, Cyclists, and Vehicles classes.
It shows that even without any supervision the backbone's embedding encapsulates object awareness, where objects from the same category are clustered together.
We provide more examples at the end of the supplemental material.

\subsection{Quantitative results}
We present additional in-domain and out-of-domain quantitative results. 
Specifically, we provide 3D detection results on the KITTI dataset~\cite{kitti1}, including various difficulty levels. 
Additionally, we present 3D detection results for the Waymo dataset~\cite{waymo} on the Level-1 and Level-2 difficulty levels. 
Finally, we provide all the experiments related to data efficiency with and without frozen features.

\subsubsection{Transfer learning on KITTI (out-of-domain).}
We start by pre-training on the entire Waymo training split. 
Following this, we perform fine-tuning with varying amounts of labeled data from KITTI's train split and report the results on the entire validation split. 
Specifically, we partition the train set into $20\%$, $50\%$, and $100\%$, resulting in approximately $0.7K$, $1.9K$, and $3.7K$ scenes, respectively.

In \tabref{tab:finetune_kitti_results_full}, we present the results for the PV-RCNN~\cite{pvrcnn} detector on the KITTI 3D detection benchmark. 
The results demonstrate that our approach consistently improves performance compared to training from scratch. 
Notably, the improvement is more significant when a smaller amount of labeled data is available. 
When using all training examples ($100\%$), we achieve results on par with those presented in \cite{proposalcontrast} and \cite{ALSO}.

\begin{table*}
    \scriptsize
    \setlength\tabcolsep{6pt} % default value: 6pt
    \centering
    \begin{tabular}{ c | c | c | c c c | c c c | c c c }
         \toprule
            \multirow{2}{*}{Labels} & \multirow{2}{*}{Method} & \multicolumn{1}{c|}{mAP} & \multicolumn{3}{c|}{\em Car}  & \multicolumn{3}{c}{\em Pedestrian}  & \multicolumn{3}{|c}{\em Cyclist} \\
             & & \multicolumn{1}{c|}{Mod.} & \multicolumn{1}{c}{Easy} & \multicolumn{1}{c}{Mod.} & \multicolumn{1}{c|}{Hard} & \multicolumn{1}{c}{Easy} & \multicolumn{1}{c}{Mod.} & \multicolumn{1}{c|}{Hard} & \multicolumn{1}{c}{Easy} & \multicolumn{1}{c}{Mod.} & \multicolumn{1}{c}{Hard} \\  
        \midrule
            \multirow{3}{*}{20\%} & PV-RCNN & 66.71 & 91.81 & 82.52 & 80.11 & 58.78 & 53.33 & 47.61 & 86.74 & 64.28 & 59.53 \\
            & ProposalContrast & 68.13 & 91.96 & 82.65 & 80.15 & 62.58 & 55.05 & 50.06 & 88.58 & 66.68 & 62.32 \\
            & PatchContrast (Ours) & \textbf{70.75} & 91.81 & 82.63 & 81.83 & 65.95 & 57.77 & 52.94 & 90.54 & 71.84 & 67.25 \\ % Ours (ckp10, best)
        \midrule
            \multirow{3}{*}{50\%} & PV-RCNN & 69.63 & 91.77 & 82.68 & 81.9 & 63.70 & 57.10 & 52.77 & 89.77 & 69.12 & 64.61 \\
            & ProposalContrast & 71.76 & 92.29 & 82.92 & 82.09 & 65.82 & 59.92 & 55.06 & 91.87 & 72.45 & 67.53 \\
            & PatchContrast (Ours) & \textbf{72.39} & 91.78 & 84.47 & 82.23 & 68.21 & 60.76 & 54.84 & 90.59 & 71.94 & 67.37 \\ % Ours (ckp15, best)
        \midrule
            \multirow{7}{*}{100\%} & PV-RCNN & 70.57 & - & 84.50 & - & - & 57.06 & - & - & 70.14 & - \\
            & GCC-3D & 71.26 & - & - & - & - & - & - & - & - & - \\
            & STRL & 71.46 & - & 84.70 & - & - & 57.80 & - & - & 71.88 & - \\
            & PointContrast & 71.55 & 91.40 & 84.18 & 82.25 & 65.73 & 57.74 & 52.46 & 91.47 & 72.72 & 67.95 \\
            & ProposalContrast & 72.92 & 92.45 & 84.72 & 82.47 & 68.43 & 60.36 & 55.01 & 92.77 & 73.69 & 69.51 \\
            & ALSO & 72.96 & - & 84.68 & - & - & 60.16 & - & - & 74.04 & - \\
            & PatchContrast (Ours) & \textbf{72.97} & 92.08 & 84.67 & 82.35 & 66.95 & 59.92 & 54.43 & 91.83 & 74.33 & 69.83 \\ %  Ours (best epoch)
         \bottomrule
    \end{tabular}
    \caption{
    \textbf{Transfer learning on KITTI.} 
    Performance comparison on the KITTI validation set. 
    The improvement is more significant when a smaller amount of labeled data is available.
    }
    \label{tab:finetune_kitti_results_full}
\end{table*}

% \begin{table}
%     \scriptsize
%     \setlength\tabcolsep{6pt} % default value: 6pt
%     \centering
%     \begin{tabular}{ c | c | c | c c c }
%          \toprule
%             Labels & Method & mAP & {\em Car} & {\em Pedestrian} & {\em Cyclist} \\
%         \midrule
%             \multirow{3}{*}{20\%} & Scratch & 66.71 & 82.52 & 53.33 & 64.28 \\
%             & PropCont~\cite{proposalcontrast} & 68.13  & 82.65 & 55.05 &  66.68 \\
%             & PatchContrast (Ours) & \textbf{70.75} & 82.63  & 57.77 &  71.84 \\ % Ours (ckp10, best)
%         \midrule
%             \multirow{3}{*}{50\%} & Scratch & 69.63 & 82.68  & 57.10  & 69.12 \\
%             & PropCont~\cite{proposalcontrast} & 71.76 & 82.92 &  59.92  & 72.45 \\
%             & PatchContrast (Ours) & \textbf{72.39} & 84.47  & 60.76 &  71.94 \\ % Ours (ckp15, best)
%         \midrule
%             \multirow{3}{*}{100\%} & Scratch & 70.57  & 84.50 & 57.06 & 70.14 \\
%             % & GCC-3D~\cite{GCC-3D} & 71.26 & - & - & -  \\
%             % & STRL~\cite{STRL} & 71.46  & 84.70 & 57.80 & 71.88  \\
%             % & PointContrast\cite{pointcontrast} & 71.55  & 84.18 & 57.74 &  72.72  \\
%             & PropCont~\cite{proposalcontrast} & 72.92  & 84.72 &  60.36 &  73.69 \\
%             % & ALSO~\cite{ALSO} & 72.96  & 84.68  & 60.16  & 74.04  \\
%             & PatchContrast (Ours) & \textbf{72.97}  & 84.67 & 59.92 & 74.33 \\ %  Ours (best epoch)
%          \bottomrule
%     \end{tabular}
    
% \end{table}

\subsubsection{Object detection on Waymo dataset (in-domain)}
As a complement to the results presented in \tabref{tab:finetune_waymo_results}, where we reported results for Waymo Level-2 difficulty, we now provide results for the Level-1 difficulty aswell. 
To ensure consistency with the common protocol~\cite{openpcdet2020}, we conducted fine-tuning with $20\%$ of labeled examples (approximately 31.6K scenes) from the training set, training for 30 epochs, and subsequently evaluated on the validation set.

The results are detailed in \tabref{tab:finetune_waymo_results_full} and are compared to other state-of-the-art pre-training methods, namely GCC-3D~\cite{GCC-3D} and ProposalContrast~\cite{proposalcontrast}. 
Notably, GCC-3D and ProposalContrast report results for Level-2 and are presented in this study for reference.

Our findings demonstrate that our approach not only improves performance over the baseline (training from scratch) but also outperforms other methods. 
It's worth noting that our approach achieves these improvements while exclusively utilizing the training split data for pre-training, in contrast to other methods that also leverage the validation split.

\begin{table*}[h]
    \scriptsize
    \setlength\tabcolsep{4pt} % default value: 6pt
    \centering
    \begin{tabular}{l | c c | c c | c c | c c | c c | c c | c c | c c}
         \toprule
            \multirow{2}{*}{Method} & \multicolumn{2}{c|}{Overall (L1)} & \multicolumn{2}{c|}{Overall (L2)} & \multicolumn{2}{c|}{{\em Vehicle} (L1)} & \multicolumn{2}{c|}{{\em Vehicle} (L2)} & \multicolumn{2}{c|}{{\em Ped.} (L1)} & \multicolumn{2}{c|}{{\em Ped.} (L2)} & \multicolumn{2}{c|}{{\em Cyc.} (L1)} & \multicolumn{2}{c}{{\em Cyc.} (L2)} \\
             & \multicolumn{1}{c}{mAP} & \multicolumn{1}{c|}{mAPH} & \multicolumn{1}{c}{mAP} & \multicolumn{1}{c|}{mAPH} & \multicolumn{1}{c}{AP} & \multicolumn{1}{c|}{APH} & \multicolumn{1}{c}{AP} & \multicolumn{1}{c|}{APH} & \multicolumn{1}{c}{AP} & \multicolumn{1}{c|}{APH} & \multicolumn{1}{c}{AP} & \multicolumn{1}{c|}{APH} & \multicolumn{1}{c}{AP} & \multicolumn{1}{c|}{APH} & \multicolumn{1}{c}{AP} & \multicolumn{1}{c}{APH}\\ 
        \midrule
            PV-RCNN & 71.09 & 66.74 & 64.84 & 60.86 & 75.41 & 74.74 & 67.44 & 66.80 & 71.98 & 61.24 & 63.70 & 53.95 & 65.88 & 64.25 & 63.39 & 61.82 \\
            GCC-3D & - & - & 61.30 & 58.18 & - & - & 65.65 & 65.10 & - & - & 55.54 & 48.02 & - & - & 62.72 & 61.43 \\
            ProposalContrast & - & - & 62.62 & 59.28 & - & - & 66.04 & 65.47 & - & - & 57.58 & 49.51 & - & - & 64.23 & 62.86  \\
            PatchContrast (Ours) & \textbf{74.59} & \textbf{70.46} & \textbf{67.91} & \textbf{64.14} & 76.90 & 76.29 & 68.40 & 67.84 & 75.51 & 65.41 & 66.62 & 57.48 & 71.35 & 69.68 & 68.72 & 67.11 \\
        \midrule
            CenterPoint & 72.66 & 69.99 & 66.48 & 64.01 & 72.76 & 72.23 & 64.91 & 64.42 & 74.19 & 67.96 & 66.03 & 60.34 & 71.04 & 69.79 & 68.49 & 67.28 \\
            GCC-3D & - & - & 65.29 & 62.79 & - & - & 63.97 & 63.47 & - & - & 64.23 & 58.47 & - & - & 67.68 & 66.44 \\
            ProposalContrast & - & - & 66.42 & 63.85 & - & - & 64.94 & 64.42 & - & - & 66.13 & 60.11 & - & - & 68.19 & 67.01  \\
            PatchContrast (Ours) & \textbf{73.22} & \textbf{70.58} & \textbf{67.02} & \textbf{64.57} & 72.84 & 72.31 & 64.73 & 64.25 & 74.99 & 68.85 & 67.10 & 61.45 & 71.84 & 70.59 & 69.22 & 68.01 \\
         \bottomrule
    \end{tabular}
    \caption{\textbf{3D detection results on Waymo.} 
    Performance comparison of Level-1 (L1) and Level-2 (L2) on the Waymo validation set, trained on 20\% of the Waymo train set, demonstrates that we outperform previous state-of-the-art methods.
    }
    \label{tab:finetune_waymo_results_full}
\end{table*}

\subsubsection{Data efficiency on Waymo dataset (in-domain)}
In this experiment, we assess the performance of our pre-trained backbones in a data-efficient setting, where we employ different amounts of labeled data.
Specifically, we divided the Waymo training set into two groups comprising $399$ sequences, equal to about $80K$ frames. 
The first $399$ sequences were used for pre-training, while various amounts of labeled data from the remaining $399$ sequences were utilized for fine-tuning.

We conducted fine-tuning for CenterPoints~\cite{yin2021center}, PV-RCNN~\cite{pvrcnn}, and SECOND~\cite{second}, using $1\%$, $5\%$, and $10\%$ of Waymo's~\cite{waymo} train set for $12$ epochs, followed by evaluation on the validation set. 
Each experiment was repeated $3$ times for consistency.

In \tabref{tab:waymo_label_efficiency_full}, we provide results for both Level-1 (L1) and Level-2 (L2), reporting both the averages and the standard deviations for each experiment. 
The results demonstrate that our pre-trained framework delivers substantial benefits, particularly when working with limited labeled data.

\begin{table*}
    \scriptsize
    \setlength\tabcolsep{3pt} % default value: 6pt
    \centering
    \begin{tabular}{c | c | c c | c c | c c | c c | c c | c c | c c | c c}
         \toprule
            \multirow{2}{*}{Labels} & \multirow{2}{*}{Method} & \multicolumn{2}{c|}{{\em Vehicle} (L1)} & \multicolumn{2}{c|}{{\em Vehicle} (L2)} & \multicolumn{2}{c|}{{\em Ped.} (L1)} & \multicolumn{2}{c|}{{\em Ped.} (L2)} & \multicolumn{2}{c|}{{\em Cyc.} (L1)} & \multicolumn{2}{c|}{{\em Cyc.} (L2)} & \multicolumn{2}{c|}{Average and std (L1)} & \multicolumn{2}{c}{Average and std  (L2)} \\
             & &  \multicolumn{1}{c}{AP} & \multicolumn{1}{c|}{APH} & \multicolumn{1}{c}{AP} & \multicolumn{1}{c|}{APH} & \multicolumn{1}{c}{AP} & \multicolumn{1}{c|}{APH} & \multicolumn{1}{c}{AP} & \multicolumn{1}{c|}{APH} & \multicolumn{1}{c}{AP} & \multicolumn{1}{c|}{APH} & \multicolumn{1}{c}{AP} & \multicolumn{1}{c|}{APH} & \multicolumn{1}{c}{AP} & \multicolumn{1}{c|}{APH} & \multicolumn{1}{c}{AP} & \multicolumn{1}{c}{APH}\\ 
        \midrule
            \multirow{21}{*}{1\%} & \multirow{3}{*}{CenterPoint} & 33.50 & 32.82 & 28.81 & 28.22 & 38.16 & 28.80 & 32.52 & 24.52 & 22.45 & 21.32 & 21.59 & 20.50 & \multirow{3}{*}{30.84$\pm$0.76} & \multirow{3}{*}{26.88$\pm$1.22} & \multirow{3}{*}{27.17$\pm$0.72} & \multirow{3}{*}{23.75$\pm$1.11} \\
             & & 31.89 & 31.23 & 27.42 & 26.85 & 37.49 & 28.38 & 31.95 & 24.17 & 24.16 & 22.95 & 23.23 & 22.07 & & & & \\
             & & 30.92 & 30.28 & 26.57 & 26.01 & 38.12 & 26.67 & 32.40 & 22.66 & 20.88 & 19.46 & 20.08 & 18.72 & & & &  \\
        \cmidrule{2-18}
              & \multirow{3}{*}{Ours} & 44.48 & 43.75 & 38.47 & 37.84 & 46.31 & 34.83 & 39.76 & 29.88 & 25.80 & 24.10 & 24.82 & 23.18 & \textbf{\multirow{3}{*}{38.84$\pm$0.17}} & \textbf{\multirow{3}{*}{34.11$\pm$0.35}} & \textbf{\multirow{3}{*}{34.43$\pm$0.18}} & \textbf{\multirow{3}{*}{30.29$\pm$0.31}} \\
              & & 43.01 & 42.28 & 37.17 & 36.54 & 42.64 & 31.57 & 36.61 & 27.06 & 31.32 & 29.30 & 30.12 & 28.18 & & & &  \\
              & & 42.81 & 42.07 & 37.00 & 36.36 & 43.82 & 31.73 & 37.71 & 27.28 & 29.33 & 27.34 & 28.21 & 26.29 & & & &  \\
        \cmidrule{2-18}
              & \multirow{3}{*}{PV-RCNN} & 48.47 & 38.75 & 41.89 & 33.50 & 29.39 & 14.87 & 24.53 & 12.41 & 17.73 & 9.60 & 17.05 & 9.23 & \multirow{3}{*}{30.83$\pm$2.41} & \multirow{3}{*}{21.1$\pm$1.26} & \multirow{3}{*}{26.87$\pm$2.18} & \multirow{3}{*}{18.38$\pm$1.16} \\
              & & 47.72 & 40.54 & 41.20 & 35.01 & 24.85 & 12.58 & 20.70 & 10.48 & 11.65 & 6.43 & 11.21 & 6.19 & & & &  \\
              & & 49.47 & 41.60 & 42.78 & 36.05 & 31.03 & 15.50 & 25.93 & 12.96 & 17.17 & 10.02 & 16.51 & 9.63 & & & &  \\
        \cmidrule{2-18}
              & \multirow{3}{*}{Ours} & 54.84 & 38.40 & 47.58 & 33.30 & 39.50 & 18.93 & 33.25 & 15.93 & 31.77 & 18.34 & 30.55 & 17.63 & \textbf{\multirow{3}{*}{43.45$\pm$1.99}} & \textbf{\multirow{3}{*}{28.50$\pm$3.16}} & \textbf{\multirow{3}{*}{38.40$\pm$1.81}} & \textbf{\multirow{3}{*}{25.18$\pm$2.79}} \\
              & & 55.68 & 45.46 & 48.31 & 39.41 & 40.35 & 20.51 & 33.98 & 17.27 & 31.74 & 20.34 & 30.52 & 19.56 & & & &  \\
              & & 57.78 & 52.61 & 50.21 & 45.67 & 44.37 & 21.28 & 37.50 & 17.99 & 35.04 & 20.68 & 33.69 & 19.89 & & & &  \\
        \cmidrule{2-18}
             & \multirow{3}{*}{SECOND} & 40.39 & 37.19 & 34.80 & 32.04 & 29.67 & 14.49 & 25.00 & 12.21 & 2.52 & 1.18 & 2.42 & 1.14 & \multirow{3}{*}{23.69 $\pm$0.64} & \multirow{3}{*}{17.28 $\pm$0.58} & \multirow{3}{*}{20.31 $\pm$0.55} & \multirow{3}{*}{14.83 $\pm$0.50} \\
             & & 38.08 & 34.02 & 32.78 & 29.28 & 27.77 & 14.37 & 23.35 & 12.08 & 3.06 & 1.44 & 2.94 & 1.39 & & & &  \\
             & & 39.73 & 36.28 & 34.21 & 31.22 & 29.19 & 15.11 & 24.59 & 12.73 & 2.76 & 1.46 & 2.66 & 1.40 & & & &  \\
        \cmidrule{2-18}
             & \multirow{3}{*}{Ours} & 46.42 & 42.82 & 40.19 & 37.07 & 36.23 & 18.13 & 30.70 & 15.36 & 22.38 & 12.91 & 21.52 & 12.41 & \textbf{\multirow{3}{*}{34.21 $\pm$3.37}} & \textbf{\multirow{3}{*}{24.43 $\pm$2.75}} & \textbf{\multirow{3}{*}{30.05 $\pm$3.14}} & \textbf{\multirow{3}{*}{21.42 $\pm$2.56}}  \\
             & & 44.57 & 40.44 & 38.51 & 34.93 & 33.27 & 17.30 & 28.15 & 14.63 & 13.67 & 7.01 & 13.14 & 6.74 & & & &  \\
             & & 46.59 & 43.52 & 40.31 & 37.65 & 37.58 & 20.03 & 31.85 & 16.97 & 27.13 & 17.68 & 26.09 & 17.01 & & & &  \\
        \midrule
            \multirow{21}{*}{5\%} & \multirow{3}{*}{CenterPoint} & 56.65 & 55.96 & 49.36 & 48.75 & 55.43 & 47.81 & 48.2 & 41.51 & 53.57 & 52.18 & 51.52 & 50.18 & \multirow{3}{*}{55.28$\pm$0.60} & \multirow{3}{*}{52.08$\pm$0.52} & \multirow{3}{*}{49.74$\pm$0.55} & \multirow{3}{*}{46.90$\pm$0.48} \\
            & & 57.59 & 56.85 & 50.19 & 49.55 & 55.86 & 48.16 & 48.59 & 41.83 & 54.26 & 52.89 & 52.18 & 50.87 & & & &  \\
            & & 57.34 & 56.62 & 49.95 & 49.32 & 54.32 & 47.06 & 47.23 & 40.86 & 52.48 & 51.18 & 50.48 & 49.22 & & & &  \\
        \cmidrule{2-18}
             & \multirow{3}{*}{Ours} & 59.18 & 58.48 & 51.69 & 51.08 & 55.24 & 47.54 & 48.17 & 41.39 & 57.39 & 55.89 & 55.21 & 53.76 & \textbf{\multirow{3}{*}{56.98$\pm$0.28}} & \textbf{\multirow{3}{*}{53.7$\pm$0.27}} & \textbf{\multirow{3}{*}{51.40$\pm$0.27}} & \textbf{\multirow{3}{*}{48.48$\pm$0.26}} \\
             & & 59.28 & 58.55 & 51.73 & 51.09 & 54.96 & 47.23 & 47.93 & 41.12 & 55.89 & 54.50 & 53.77 & 52.44 & & & &  \\
             & & 59.51 & 58.81 & 51.96 & 51.34 & 55.31 & 47.61 & 48.26 & 41.48 & 56.04 & 54.69 & 53.93 & 52.62 & & & &  \\
        \cmidrule{2-18}
              &  \multirow{3}{*}{PV-RCNN} & 67.28 & 65.97 & 58.88 & 57.71 & 56.89 & 28.89 & 48.62 & 24.69 & 50.89 & 29.55 & 48.94 & 28.42 & \multirow{3}{*}{58.16$\pm$0.18} & \multirow{3}{*}{40.76$\pm$0.70} & \multirow{3}{*}{51.97$\pm$0.16} & \multirow{3}{*}{36.27$\pm$0.65} \\
              & & 67.31 & 66.04 & 58.92 & 57.79 & 56.8 & 28.26 & 48.55 & 24.16 & 49.92 & 25.94 & 48.01 & 24.95 & & & &  \\
              & & 67.56 & 66.31 & 59.16 & 58.05 & 56.98 & 28.24 & 48.70 & 24.14 & 49.80 & 27.61 & 47.90 & 26.55 & & & &  \\
        \cmidrule{2-18}
              & \multirow{3}{*}{Ours} & 68.72 & 67.69 & 60.28 & 59.35 & 58.49 & 29.88 & 50.14 & 25.61 & 55.66 & 32.62 & 53.54 & 31.38 & \textbf{\multirow{3}{*}{60.97$\pm$0.06}} & \textbf{\multirow{3}{*}{42.81$\pm$0.54}} & \textbf{\multirow{3}{*}{54.66$\pm$0.06}} & \textbf{\multirow{3}{*}{38.21$\pm$0.52}} \\
              & & 68.70 & 67.73 & 60.28 & 59.41 & 59.09 & 29.92 & 50.67 & 25.65 & 54.95 & 29.36 & 52.85 & 28.23 & & & &  \\
              & & 68.50 & 67.43 & 60.05 & 59.09 & 58.86 & 30.04 & 50.50 & 25.78 & 55.75 & 30.58 & 53.62 & 29.42 & & & &  \\
        \cmidrule{2-18}
             & \multirow{3}{*}{SECOND} & 56.30 & 55.22 & 48.93 & 47.99 & 46.67 & 24.82 & 39.84 & 21.17 & 34.64 & 17.60 & 33.32 & 16.93 & \multirow{3}{*}{45.64$\pm$0.20} & \multirow{3}{*}{32.44$\pm$0.31} & \multirow{3}{*}{40.49$\pm$0.18} & \multirow{3}{*}{28.65$\pm$0.29} \\
             & & 55.33 & 54.23 & 48.10 & 47.15 & 47.03 & 23.71 & 40.12 & 20.23 & 34.17 & 20.10 & 32.86 & 19.33 & & & &  \\
             & & 55.53 & 54.35 & 48.27 & 47.23 & 46.75 & 23.33 & 39.91 & 19.91 & 34.38 & 18.58 & 33.07 & 17.87 & & & &  \\
        \cmidrule{2-18}
             & \multirow{3}{*}{Ours} & 59.44 & 58.45 & 51.84 & 50.97 & 50.22 & 24.72 & 43.10 & 21.22 & 42.83 & 20.28 & 41.20 & 19.50 & \textbf{\multirow{3}{*}{50.42$\pm$0.58}} & \textbf{\multirow{3}{*}{34.81$\pm$0.59}} & \textbf{\multirow{3}{*}{45.00$\pm$0.54}} & \textbf{\multirow{3}{*}{30.90$\pm$0.55}} \\
             & & 58.93 & 57.95 & 51.39 & 50.53 & 49.67 & 25.07 & 42.62 & 21.51 & 43.40 & 23.45 & 41.74 & 22.56 & & & &  \\
             & & 58.23 & 57.13 & 50.79 & 49.82 & 49.72 & 23.91 & 42.65 & 20.52 & 41.31 & 22.35 & 39.73 & 21.50 & & & &  \\
        \midrule
            \multirow{21}{*}{10\%} & \multirow{3}{*}{CenterPoint} & 62.16 & 61.49 & 54.44 & 53.85 & 62.08 & 55.14 & 54.45 & 48.27 & 61.36 & 59.84 & 59.03 & 57.57 & \multirow{3}{*}{61.62$\pm$0.37} & \multirow{3}{*}{58.58$\pm$0.38} & \multirow{3}{*}{55.74$\pm$0.37} & \multirow{3}{*}{53.00$\pm$0.37} \\
            & & 62.41 & 61.75 & 54.62 & 54.04 & 61.96 & 54.97 & 54.42 & 48.18 & 61.04 & 59.60 & 58.74 & 57.35 & & & &  \\
            & & 62.31 & 61.63 & 54.55 & 53.96 & 61.07 & 54.06 & 53.48 & 47.25 & 60.21 & 58.71 & 57.92 & 56.49 & & & &  \\
        \cmidrule{2-18}
             & \multirow{3}{*}{Ours} & 63.04 & 62.36 & 55.24 & 54.65 & 62.07 & 54.78 & 54.40 & 47.92 & 61.49 & 60.09 & 59.17 & 57.82 & \textbf{\multirow{3}{*}{62.15$\pm$0.06}} & \textbf{\multirow{3}{*}{59.01$\pm$0.07}} & \textbf{\multirow{3}{*}{56.24$\pm$0.05}} & \textbf{\multirow{3}{*}{53.41$\pm$0.06}} \\
             & & 63.28 & 62.62 & 55.48 & 54.89 & 61.95 & 54.58 & 54.34 & 47.78 & 61.03 & 59.62 & 58.73 & 57.37 & & & &  \\
             & & 63.57 & 62.91 & 55.72 & 55.14 & 61.66 & 54.26 & 54.12 & 47.53 & 61.24 & 59.82 & 58.94 & 57.57 & & & &  \\
        \cmidrule{2-18}
              & \multirow{3}{*}{PV-RCNN} & 70.36 & 69.45 & 61.79 & 60.98 & 61.92 & 30.42 & 53.26 & 26.17 & 56.95 & 31.44 & 54.77 & 30.24 & \multirow{3}{*}{62.84$\pm$0.57} & \multirow{3}{*}{43.4$\pm$0.35} & \multirow{3}{*}{56.38$\pm$0.55} & \multirow{3}{*}{38.77$\pm$0.34} \\
              & & 70.05 & 69.11 & 61.52 & 60.66 & 60.74 & 30.36 & 52.14 & 26.06 & 55.77 & 30.62 & 53.63 & 29.45 & & & &  \\
              & & 70.55 & 69.65 & 61.99 & 61.18 & 61.46 & 30.68 & 52.82 & 26.37 & 57.74 & 28.91 & 55.54 & 27.81 & & & &  \\
        \cmidrule{2-18}
              & \multirow{3}{*}{Ours} & 70.81 & 70.00 & 62.26 & 61.52 & 62.35 & 31.08 & 53.70 & 26.77 & 59.38 & 32.74 & 57.12 & 31.50 & \textbf{\multirow{3}{*}{64.13$\pm$0.04}} & \textbf{\multirow{3}{*}{44.08$\pm$0.53}} & \textbf{\multirow{3}{*}{57.65$\pm$0.04}} & \textbf{\multirow{3}{*}{39.43$\pm$0.48}} \\
              & & 70.79 & 69.98 & 62.24 & 61.51 & 62.56 & 32.06 & 53.89 & 27.61 & 58.94 & 30.27 & 56.71 & 29.12 & & & &  \\
              & & 70.91 & 70.09 & 62.39 & 61.64 & 62.68 & 29.81 & 53.97 & 25.68 & 58.78 & 30.72 & 56.55 & 29.56 & & & &  \\
        \cmidrule{2-18}
             & \multirow{3}{*}{SECOND} & 61.03 & 60.04 & 53.32 & 52.45 & 53.15 & 27.24 & 45.71 & 23.42 & 43.74 & 25.07 & 42.07 & 24.11 & \multirow{3}{*}{51.67$\pm$1.15} & \multirow{3}{*}{36.65$\pm$1.02} & \multirow{3}{*}{46.12$\pm$1.08} & \multirow{3}{*}{32.56$\pm$0.97} \\
             & & 60.17 & 59.22 & 52.52 & 51.68 & 51.36 & 26.52 & 44.10 & 22.77 & 39.68 & 20.75 & 38.17 & 19.96 & & & &  \\
             & & 60.80 & 59.89 & 53.08 & 52.27 & 52.55 & 27.33 & 45.13 & 23.47 & 42.59 & 23.81 & 40.96 & 22.90 & & & &  \\
        \cmidrule{2-18}
             & \multirow{3}{*}{Ours} & 62.67 & 61.78 & 54.86 & 54.06 & 54.74 & 27.67 & 47.23 & 23.88 & 47.56 & 26.57 & 45.76 & 25.56 & \textbf{\multirow{3}{*}{54.58$\pm$0.44}} & \textbf{\multirow{3}{*}{38.32$\pm$0.63}} & \textbf{\multirow{3}{*}{48.89$\pm$0.41}} & \textbf{\multirow{3}{*}{34.17$\pm$0.58}} \\
             & & 62.90 & 62.08 & 55.03 & 54.31 & 54.33 & 27.57 & 46.85 & 23.76 & 46.70 & 26.48 & 44.93 & 25.47 & & & &  \\
             & & 62.12 & 61.24 & 54.37 & 53.59 & 53.95 & 26.52 & 46.50 & 22.87 & 46.27 & 25.01 & 44.51 & 24.06 & & & &  \\
         \bottomrule
    \end{tabular}
    \caption{
    \textbf{Data efficiency on Waymo.} 
    All of the experiments.
    }
    \label{tab:waymo_label_efficiency_full}
\end{table*}

\subsubsection{Data efficiency on Waymo with frozen features (in-domain)}
Similar to the linear classification protocol utilized in 2D image domains~\cite{simclr, moco}, we propose freezing the features of the backbone and training a detection head on top of them. 
This approach allows us to evaluate the feature embeddings before the fine-tuning process overwrites the backbone's weights.

As in the previous subsection, we partition the Waymo training set into two groups consisting of $399$ sequences. 
The first $399$ sequences are used for pre-training, and different amounts of labeled data from the remaining $399$ sequences are employed for fine-tuning. We subsequently evaluate the model on the validation set.

Results for both Level-1 and Level-2 difficulties of Waymo, using different percentages of labeled data, are reported in \tabref{tab:waymo_label_efficiency_frozen_bb_full}.
Each experiment was conducted $3$ times for consistency, and we provided both the means and standard deviations for each experiment.
For the detector, we adopt CenterPoint~\cite{yin2021center} and employ the 1x scheduler (12 epochs).
The results demonstrate that our embedding captures meaningful information about 3D point cloud scenes without any fine-tuning while outperforming the previous state-of-the-art method.

\begin{table*}
    \scriptsize
    \setlength\tabcolsep{3pt} % default value: 6pt
    \centering
    \begin{tabular}{c | c | c c | c c | c c | c c | c c | c c | c c | c c}
         \toprule
            \multirow{2}{*}{Labels} & \multirow{2}{*}{Method} & \multicolumn{2}{c|}{{\em Vehicle} (L1)} & \multicolumn{2}{c|}{{\em Vehicle} (L2)} & \multicolumn{2}{c|}{{\em Ped.} (L1)} & \multicolumn{2}{c|}{{\em Ped.} (L2)} & \multicolumn{2}{c|}{{\em Cyc.} (L1)} & \multicolumn{2}{c|}{{\em Cyc.} (L2)} & \multicolumn{2}{c|}{Average and std (L1)} & \multicolumn{2}{c}{Average and std  (L2)} \\
             & &  \multicolumn{1}{c}{AP} & \multicolumn{1}{c|}{APH} & \multicolumn{1}{c}{AP} & \multicolumn{1}{c|}{APH} & \multicolumn{1}{c}{AP} & \multicolumn{1}{c|}{APH} & \multicolumn{1}{c}{AP} & \multicolumn{1}{c|}{APH} & \multicolumn{1}{c}{AP} & \multicolumn{1}{c|}{APH} & \multicolumn{1}{c}{AP} & \multicolumn{1}{c|}{APH} & \multicolumn{1}{c}{AP} & \multicolumn{1}{c|}{APH} & \multicolumn{1}{c}{AP} & \multicolumn{1}{c}{APH}\\ 
        \midrule
            \multirow{10}{*}{1\%} & \multirow{3}{*}{CenterPoint} & 9.81 & 9.24 & 8.38 & 7.89 & 9.59 & 4.88 & 7.97 & 4.06 & 0.68 & 0.41 & 0.66 & 0.39 & \multirow{3}{*}{6.27$\pm$0.47} & \multirow{3}{*}{4.47$\pm$0.36} & \multirow{3}{*}{5.30$\pm$0.39} & \multirow{3}{*}{3.80$\pm$0.30} \\
            & & 8.60 & 8.08 & 7.34 & 6.90 & 9.75 & 4.93 & 8.10 & 4.10 & 0.67 & 0.35 & 0.64 & 0.34 & & & & \\
            & & 8.11 & 7.68 & 6.93 & 6.56 & 8.43 & 4.28 & 7.00 & 3.55 & 0.76 & 0.42 & 0.73 & 0.40 & & & & \\
        \cmidrule{2-18}
             & \multirow{3}{*}{PropsalContrast} & 9.43 & 8.98 & 8.03 & 7.66 & 7.78 & 3.99 & 6.48 & 3.32 & 0.63 & 0.34 & 0.61 & 0.32 & \multirow{3}{*}{5.76$\pm$0.79} & \multirow{3}{*}{4.24$\pm$0.58} & \multirow{3}{*}{4.88$\pm$0.66} & \multirow{3}{*}{3.60$\pm$0.49} \\
             & & 7.68 & 7.31 & 6.58 & 6.26 & 6.28 & 3.13 & 5.22 & 2.60 & 0.70 & 0.32 & 0.68 & 0.30 & & & & \\
            & & 9.64 & 9.19 & 8.22 & 7.84 & 9.11 & 4.62 & 7.54 & 3.83 & 0.57 & 0.29 & 0.55 & 0.28 & & & & \\
        \cmidrule{2-18}
             & \multirow{3}{*}{Ours} & 35.66 & 34.61 & 30.69 & 29.78 & 27.48 & 14.40 & 23.24 & 12.18 & 6.21 & 3.81 & 5.97 & 3.66 & \textbf{\multirow{3}{*}{22.84$\pm$0.49}} & \textbf{\multirow{3}{*}{17.16$\pm$0.62}} & \textbf{\multirow{3}{*}{19.76$\pm$0.41}} & \textbf{\multirow{3}{*}{14.83$\pm$0.54}} \\
             & & 32.94 & 31.97 & 28.30 & 27.47 & 25.89 & 13.50 & 21.87 & 11.40 & 7.99 & 3.89 & 7.68 & 3.74  & & & & \\
            & &  34.31 & 33.25 & 29.51 & 28.59 & 27.32 & 14.42 & 23.08 & 12.18 & 7.76 & 4.61 & 7.47 & 4.43  & & & & \\
        \midrule
            \multirow{10}{*}{5\%} & \multirow{3}{*}{CenterPoint} & 29.84 & 28.75 & 25.63 & 24.69 & 17.25 & 8.94 & 14.46 & 7.49 & 4.86 & 3.43 & 4.68 & 3.30 & \multirow{3}{*}{17.14$\pm$0.53} & \multirow{3}{*}{13.46$\pm$0.41} & \multirow{3}{*}{14.75$\pm$0.46} & \multirow{3}{*}{11.60$\pm$0.36} \\
            & &  28.27 & 27.05 & 24.27 & 23.23 & 17.46 & 9.15 & 14.63 & 7.66 & 3.91 & 2.75 & 3.76 & 2.65 & & & & \\
            & &  29.84 & 28.52 & 25.62 & 24.49 & 18.25 & 9.52 & 15.28 & 7.97 & 4.56 & 3.00 & 4.39 & 2.88 & & & & \\
        \cmidrule{2-18}
             & \multirow{3}{*}{PropsalContrast} & 34.07 & 32.92 & 29.35 & 28.36 & 20.78 & 11.05 & 17.58 & 9.34 & 8.03 & 6.35 & 7.73 & 6.11 & \multirow{3}{*}{19.87$\pm$1.11} & \multirow{3}{*}{15.83$\pm$0.94} & \multirow{3}{*}{17.22$\pm$1.02} & \multirow{3}{*}{13.75$\pm$0.86} \\
             & &  31.34 & 30.21 & 26.95 & 25.98 & 19.08 & 10.05 & 16.05 & 8.45 & 5.78 & 4.42 & 5.56 & 4.25 & & & & \\
            & &  32.13 & 30.98 & 27.64 & 26.65 & 19.88 & 10.51 & 16.71 & 8.83 & 7.70 & 5.96 & 7.41 & 5.74 & & & & \\
        \cmidrule{2-18}
             & \multirow{3}{*}{Ours} & 51.00 & 49.98 & 44.23 & 43.33 & 38.82 & 22.62 & 33.18 & 19.33 & 30.75 & 27.20 & 29.57 & 26.15 & \textbf{\multirow{3}{*}{39.75$\pm$0.39}} & \textbf{\multirow{3}{*}{32.80$\pm$0.48}} & \textbf{\multirow{3}{*}{35.22$\pm$0.39}} & \textbf{\multirow{3}{*}{29.15$\pm$0.46}} \\
             & & 51.16 & 50.11 & 44.35 & 43.43 & 38.94 & 22.34 & 33.27 & 19.09 & 28.21 & 24.47 & 27.13 & 23.54  & & & & \\
            & &  51.34 & 50.39 & 44.48 & 43.65 & 38.25 & 21.95 & 32.65 & 18.73 & 29.23 & 26.13 & 28.11 & 25.13 & & & & \\
        \midrule
            \multirow{10}{*}{10\%} & \multirow{3}{*}{CenterPoint} & 35.82 & 34.78 & 30.84 & 29.94 & 21.05 & 11.39 & 17.73 & 9.59 & 7.24 & 5.29 & 6.96 & 5.08 & \multirow{3}{*}{21.55$\pm$0.18} & \multirow{3}{*}{17.19$\pm$0.07} & \multirow{3}{*}{18.69$\pm$0.17} & \multirow{3}{*}{14.92$\pm$0.08} \\
            & &  35.12 & 33.99 & 30.24 & 29.27 & 21.28 & 11.38 & 17.92 & 9.59 & 8.26 & 6.44 & 7.95 & 6.19 & & & & \\
            & & 35.76 & 34.61 & 30.80 & 29.81 & 21.86 & 11.57 & 18.43 & 9.75 & 7.60 & 5.30 & 7.31 & 5.10  & & & & \\
        \cmidrule{2-18}
             & \multirow{3}{*}{PropsalContrast} & 42.70 & 41.70 & 36.96 & 36.08 & 28.85 & 16.14 & 24.49 & 13.70 & 16.22 & 13.45 & 15.60 & 12.94 & \multirow{3}{*}{29.22$\pm$0.11} & \multirow{3}{*}{23.53$\pm$0.27} & \multirow{3}{*}{25.64$\pm$0.11} & \multirow{3}{*}{20.69$\pm$0.25} \\
             & & 42.47 & 41.35 & 36.74 & 35.77 & 29.42 & 15.99 & 24.97 & 13.58 & 15.43 & 12.38 & 14.83 & 11.90  & & & & \\
            & &  42.13 & 41.02 & 36.45 & 35.48 & 29.08 & 16.02 & 24.65 & 13.58 & 16.74 & 13.70 & 16.09 & 13.17 & & & & \\
        \cmidrule{2-18}
             & \multirow{3}{*}{Ours} & 55.07 & 54.17 & 47.88 & 47.10 & 44.28 & 26.92 & 38.04 & 23.12 & 36.85 & 33.34 & 35.44 & 32.06 & \textbf{\multirow{3}{*}{45.38$\pm$0.32}} & \textbf{\multirow{3}{*}{38.12$\pm$0.06}} & \textbf{\multirow{3}{*}{40.41$\pm$0.28}} & \textbf{\multirow{3}{*}{34.05$\pm$0.07}} \\
             & & 55.32 & 54.45 & 48.07 & 47.31 & 43.88 & 26.99 & 37.70 & 23.17 & 35.95 & 32.69 & 34.57 & 31.44  & & & & \\
            & &  55.43 & 54.52 & 48.20 & 47.40 & 45.24 & 27.19 & 38.83 & 23.33 & 36.36 & 32.77 & 34.97 & 31.51 & & & & \\
        \midrule
            \multirow{10}{*}{20\%} & \multirow{3}{*}{CenterPoint} & 39.74 & 38.70 & 34.31 & 33.40 & 23.93 & 13.16 & 20.22 & 11.11 & 10.97 & 8.63 & 10.55 & 8.30 & \multirow{3}{*}{25.00$\pm$0.72} & \multirow{3}{*}{20.13$\pm$0.52} & \multirow{3}{*}{21.80$\pm$0.67} & \multirow{3}{*}{17.58$\pm$0.49} \\
            & & 38.17 & 37.09 & 32.93 & 32.00 & 24.98 & 13.79 & 21.10 & 11.64 & 9.91 & 7.90 & 9.53 & 7.59  & & & & \\
            & & 39.94 & 38.90 & 34.50 & 33.60 & 24.81 & 13.53 & 20.97 & 11.43 & 12.56 & 9.49 & 12.08 & 9.13  & & & & \\
        \cmidrule{2-18}
             & \multirow{3}{*}{PropsalContrast} & 49.36 & 48.40 & 42.78 & 41.93 & 35.01 & 20.64 & 29.97 & 17.66 & 21.52 & 17.62 & 20.69 & 16.94 & \multirow{3}{*}{35.40$\pm$0.19} & \multirow{3}{*}{28.93$\pm$0.19} & \multirow{3}{*}{31.26$\pm$0.19} & \multirow{3}{*}{25.57$\pm$0.18} \\
             & & 48.96 & 47.92 & 42.40 & 41.50 & 34.87 & 20.14 & 29.86 & 17.24 & 22.06 & 18.24 & 21.22 & 17.54  & & & & \\
            & &  49.77 & 48.78 & 43.13 & 42.27 & 34.27 & 19.89 & 29.35 & 17.03 & 22.81 & 18.75 & 21.94 & 18.03 & & & & \\
        \cmidrule{2-18}
             & \multirow{3}{*}{Ours} & 57.27 & 56.38 & 49.89 & 49.11 & 47.75 & 30.33 & 41.08 & 26.08 & 37.65 & 34.26 & 36.20 & 32.94 & \textbf{\multirow{3}{*}{47.61$\pm$0.10}} & \textbf{\multirow{3}{*}{40.54$\pm$0.26}} & \textbf{\multirow{3}{*}{42.46$\pm$0.10}} & \textbf{\multirow{3}{*}{36.25$\pm$0.25}} \\
             & & 57.84 & 57.09 & 50.42 & 49.76 & 47.74 & 30.35 & 41.17 & 26.15 & 37.07 & 33.94 & 35.64 & 32.64  & & & & \\
            & & 57.46 & 56.71 & 50.09 & 49.43 & 47.76 & 30.43 & 41.12 & 26.19 & 37.97 & 35.34 & 36.52 & 33.99  & & & & \\
        \midrule
            \multirow{10}{*}{50\%} & \multirow{3}{*}{CenterPoint} & 47.06 & 45.93 & 40.83 & 39.84 & 29.66 & 16.74 & 25.19 & 14.21 & 16.16 & 12.77 & 15.54 & 12.28 & \multirow{3}{*}{31.27$\pm$0.30} & \multirow{3}{*}{25.44$\pm$0.29} & \multirow{3}{*}{27.47$\pm$0.28} & \multirow{3}{*}{22.40$\pm$0.27} \\
            & &  45.84 & 44.69 & 39.75 & 38.74 & 30.35 & 16.92 & 25.75 & 14.36 & 17.63 & 14.76 & 16.95 & 14.19 & & & & \\
            & &  47.06 & 45.89 & 40.84 & 39.81 & 30.47 & 17.18 & 25.88 & 14.58 & 17.17 & 14.10 & 16.51 & 13.56 & & & & \\
        \cmidrule{2-18}
             & \multirow{3}{*}{PropsalContrast} & 55.71 & 54.75 & 48.52 & 47.68 & 42.60 & 26.08 & 36.73 & 22.46 & 30.31 & 27.39 & 29.16 & 26.34 & \multirow{3}{*}{42.77$\pm$0.09} & \multirow{3}{*}{36.01$\pm$0.09} & \multirow{3}{*}{38.04$\pm$0.09} & \multirow{3}{*}{32.10$\pm$0.09} \\
             & & 55.64 & 54.71 & 48.45 & 47.63 & 42.55 & 26.22 & 36.68 & 22.58 & 29.89 & 26.78 & 28.75 & 25.75  & & & & \\
            & & 55.40 & 54.46 & 48.24 & 47.41 & 42.06 & 26.15 & 36.28 & 22.54 & 30.76 & 27.52 & 29.59 & 26.46  & & & & \\
        \cmidrule{2-18}
             & \multirow{3}{*}{Ours} & 62.30 & 61.61 & 54.59 & 53.97 & 53.01 & 35.48 & 46.03 & 30.77 & 42.71 & 39.85 & 41.08 & 38.33 & \textbf{\multirow{3}{*}{53.09$\pm$0.39}} & \textbf{\multirow{3}{*}{45.96$\pm$0.36}} & \textbf{\multirow{3}{*}{47.62$\pm$0.36}} & \textbf{\multirow{3}{*}{41.31$\pm$0.33}} \\
             & & 62.94 & 62.23 & 55.19 & 54.56 & 53.98 & 36.28 & 46.90 & 31.48 & 43.44 & 40.56 & 41.77 & 39.00  & & & & \\
            & &  62.83 & 62.14 & 55.07 & 54.45 & 53.54 & 35.78 & 46.53 & 31.05 & 43.03 & 39.72 & 41.38 & 38.20 & & & & \\
        \midrule
            \multirow{10}{*}{100\%} & \multirow{3}{*}{CenterPoint} & 49.85 & 48.73 & 43.34 & 42.35 & 32.18 & 18.23 & 27.38 & 15.50 & 20.33 & 17.04 & 19.55 & 16.38 & \multirow{3}{*}{34.16$\pm$0.16} & \multirow{3}{*}{28.00$\pm$0.01} & \multirow{3}{*}{30.12$\pm$0.14} & \multirow{3}{*}{24.75$\pm$0.01} \\
            & & 49.07 & 47.92 & 42.65 & 41.64 & 32.85 & 18.79 & 27.94 & 15.98 & 20.15 & 17.32 & 19.38 & 16.65  & & & & \\
            & &  49.39 & 48.15 & 42.94 & 41.85 & 33.02 & 18.57 & 28.09 & 15.79 & 20.57 & 17.23 & 19.78 & 16.57 & & & & \\
        \cmidrule{2-18}
             & \multirow{3}{*}{PropsalContrast} & 57.71 & 56.79 & 50.33 & 49.51 & 45.03 & 28.16 & 38.79 & 24.23 & 33.25 & 30.07 & 31.97 & 28.91 & \multirow{3}{*}{44.96$\pm$0.34} & \multirow{3}{*}{37.86$\pm$0.42} & \multirow{3}{*}{40.03$\pm$0.31} & \multirow{3}{*}{33.79$\pm$0.38} \\
             & &  57.04 & 56.09 & 49.74 & 48.90 & 44.70 & 27.17 & 38.55 & 23.41 & 32.25 & 29.37 & 31.01 & 28.23 & & & & \\
            & &  57.35 & 56.42 & 50.01 & 49.19 & 44.98 & 27.88 & 38.78 & 24.01 & 32.30 & 28.78 & 31.06 & 27.67 & & & & \\
        \cmidrule{2-18}
             & \multirow{3}{*}{Ours} & 65.36 & 64.67 & 57.50 & 56.88 & 57.13 & 38.31 & 49.82 & 33.36 & 44.98 & 41.84 & 43.26 & 40.24 & \textbf{\multirow{3}{*}{55.91$\pm$0.17}} & \textbf{\multirow{3}{*}{48.47$\pm$0.25}} & \textbf{\multirow{3}{*}{50.29$\pm$0.16}} & \textbf{\multirow{3}{*}{43.68$\pm$0.23}} \\
             & & 64.97 & 64.25 & 57.16 & 56.52 & 57.44 & 39.28 & 50.11 & 34.21 & 45.91 & 42.75 & 44.15 & 41.11  & & & & \\
            & & 65.08 & 64.38 & 57.24 & 56.62 & 56.62 & 37.98 & 49.36 & 33.06 & 45.74 & 42.79 & 43.99 & 41.15  & & & & \\
        \bottomrule
    \end{tabular}
    \caption{
    \textbf{Data efficiency on Waymo with frozen features.} 
    All of the experiments.
    }
    \label{tab:waymo_label_efficiency_frozen_bb_full}
\end{table*}

\subsection{More qualitative results}
Below we provide more qualitative results of our pre-trained backbone's embedding without any fine-tuning.

\begin{figure*}[h]
    \centering
	\includegraphics[width=.96\linewidth]{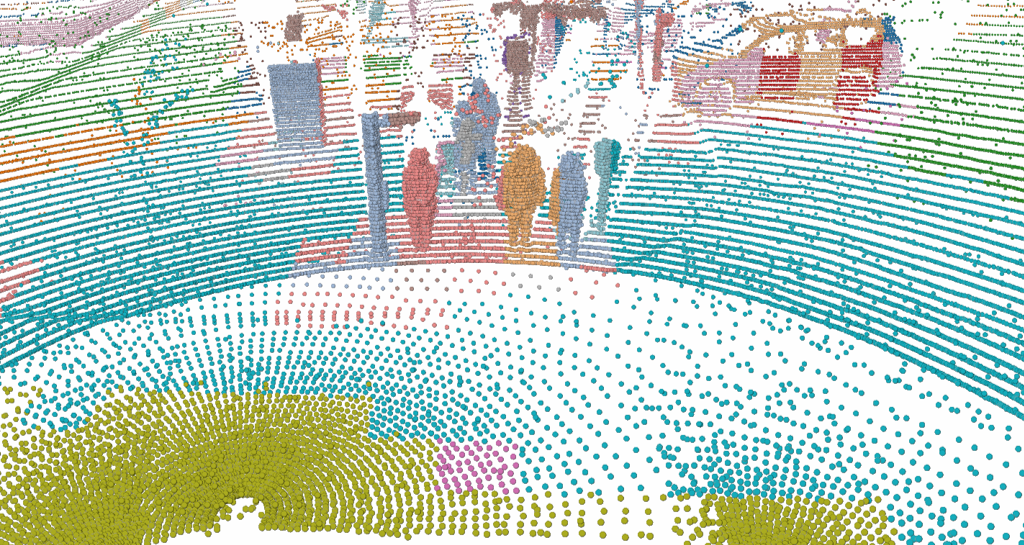}
	\caption{
    \textbf{Qualitative result.} 
    Example $1$.
    }
\end{figure*}

\begin{figure*}[h]
    \centering
	\includegraphics[width=.96\linewidth]{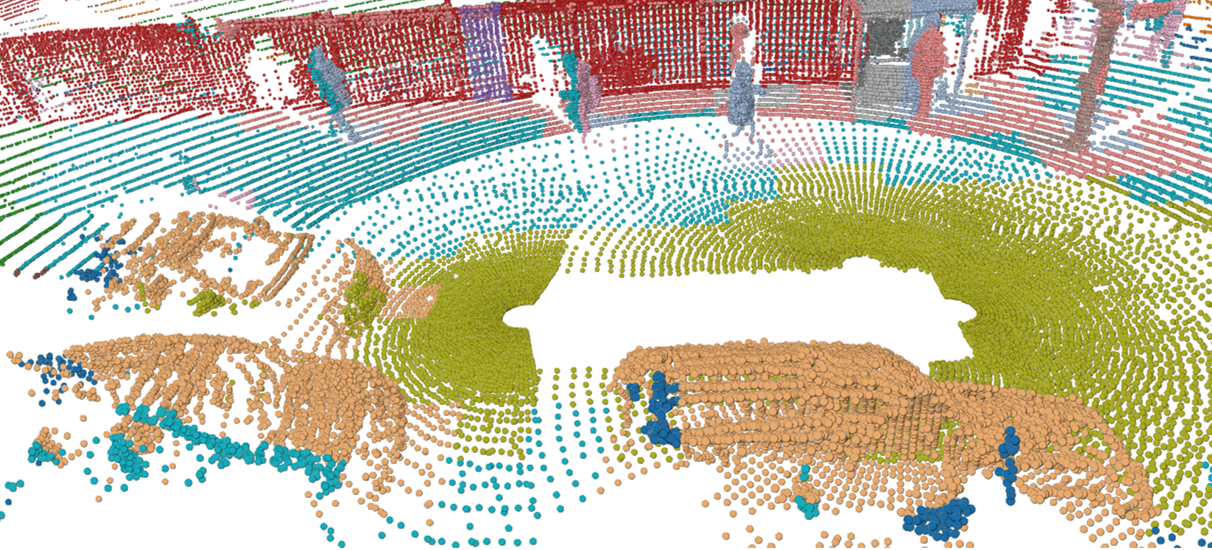}
	\caption{
    \textbf{Qualitative result.} 
    Example $2$.
    }
\end{figure*}

\begin{figure*}[h]
    \centering
	\includegraphics[width=.96\linewidth]{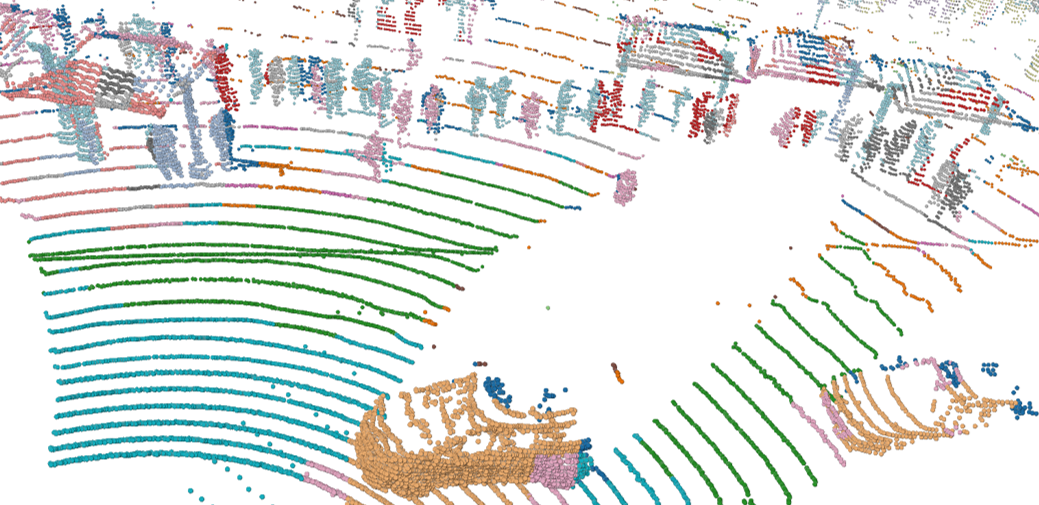}
	\caption{
    \textbf{Qualitative result.} 
    Example $3$.
    }
\end{figure*}

\begin{figure*}[h]
    \centering
	\includegraphics[width=.96\linewidth]{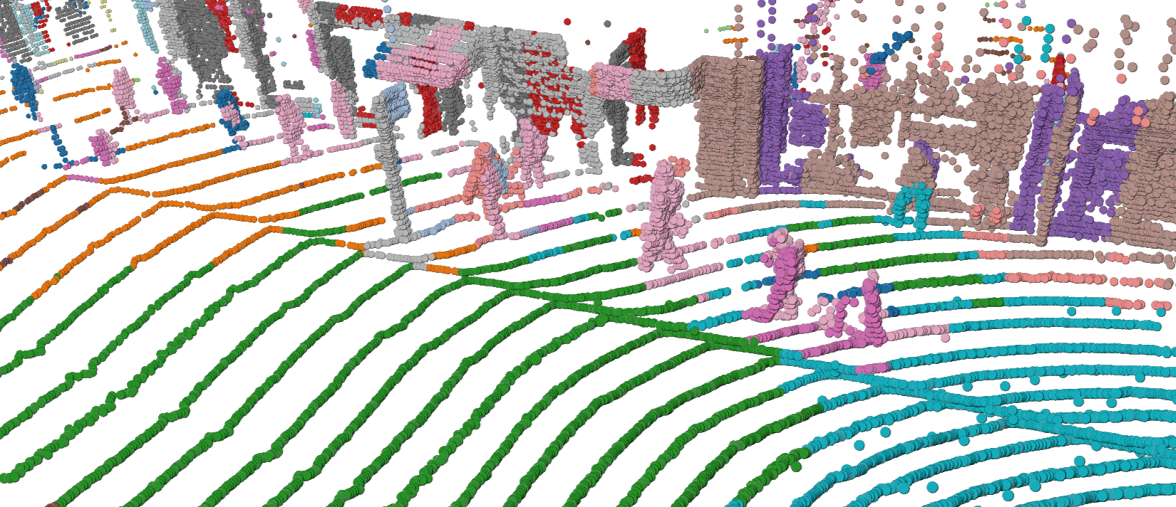}
	\caption{
    \textbf{Qualitative result.} 
    Example $4$.
    }
\end{figure*}

\begin{figure*}[h]
    \centering
	\includegraphics[width=.9\linewidth]{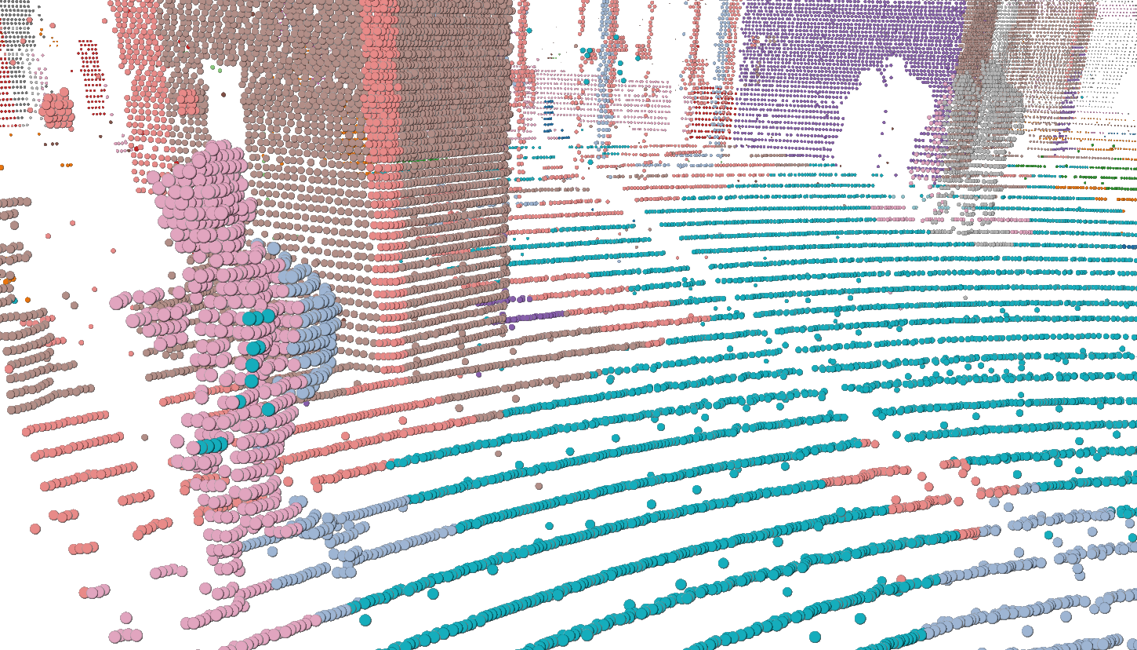}
	\caption{
    \textbf{Qualitative result.} 
    Example $5$.
    }
\end{figure*}

\begin{figure*}[h]
    \centering
	\includegraphics[width=.9\linewidth]{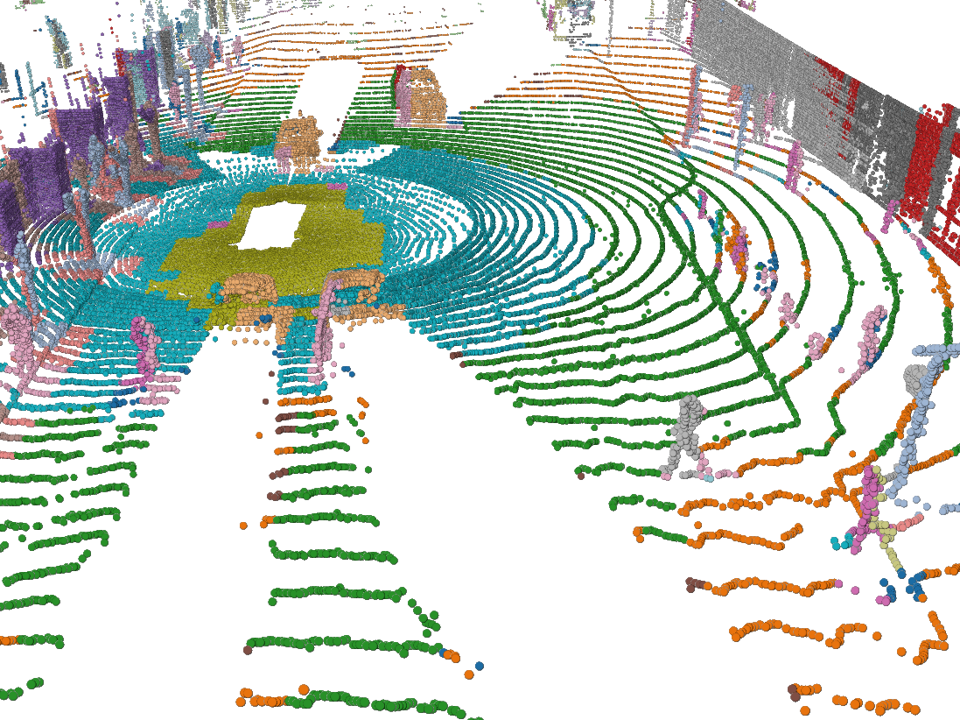}
	\caption{
    \textbf{Qualitative result.} 
    Example $6$.
    }
\end{figure*}

% \begin{figure*}[h]
%     \centering
% 	\includegraphics[width=.8\linewidth]{images/qualitative/qualitative_results_fixed_8.png}
% 	\caption{
%     \textbf{Qualitative result.} 
%     Example $7$.
%     }
% \end{figure*}

% \begin{figure*}[h]
%     \centering
% 	\includegraphics[width=.8\linewidth]{images/qualitative/qualitative_results_fixed_9.png}
% 	\caption{
%     \textbf{Qualitative result.} 
%     Example $8$.
%     }
% \end{figure*}

\begin{figure*}[h]
    \centering
	\includegraphics[width=.8\linewidth]{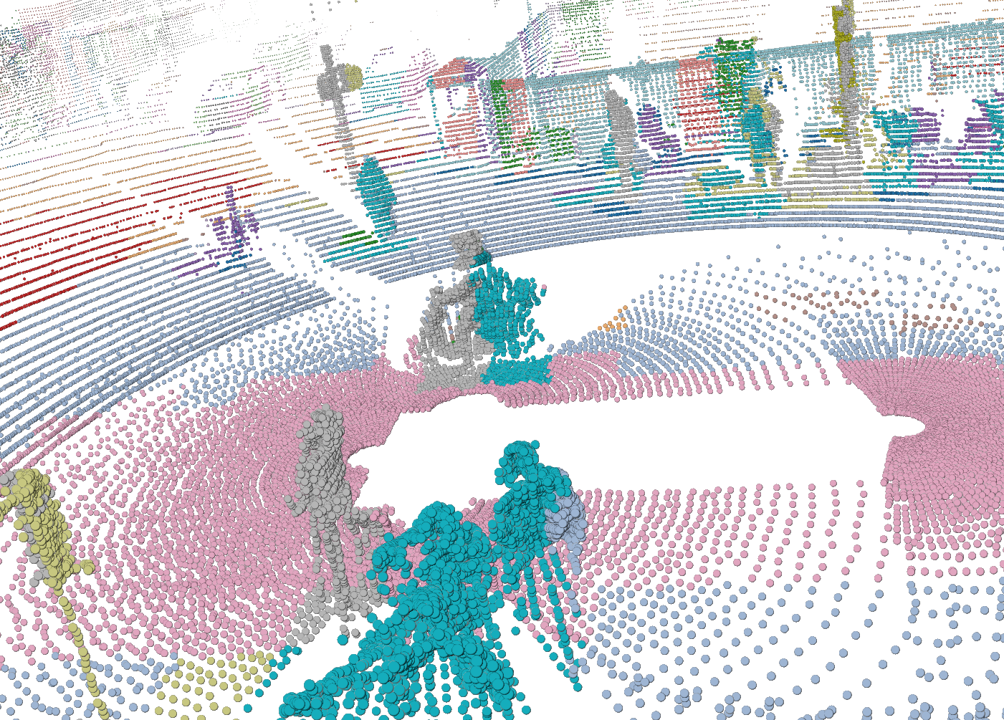}
	\caption{
    \textbf{Qualitative result.} 
    Example $9$.
    }
\end{figure*}

\begin{figure*}[h]
    \centering
	\includegraphics[width=.8\linewidth]{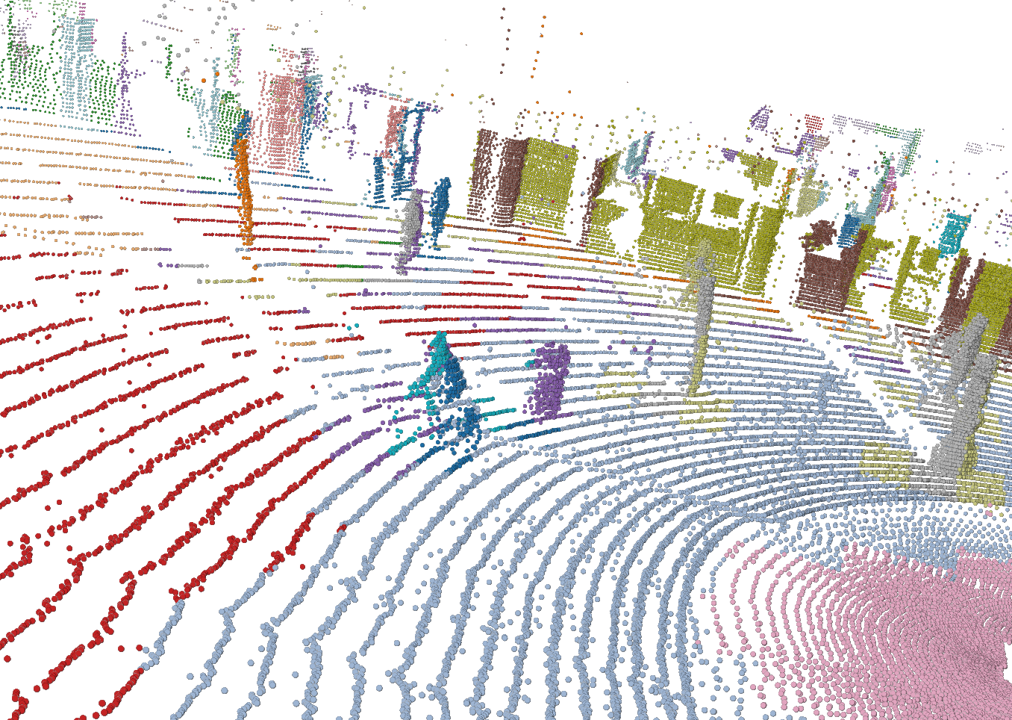}
	\caption{
    \textbf{Qualitative result.} 
    Example $10$.
    }
\end{figure*}

\end{document}